\documentclass{egpubl}
\usepackage{eg2023}
 
\ConferenceSubmission   
\CGFStandardLicense

\usepackage[T1]{fontenc}
\usepackage{dfadobe}  

\usepackage{cite}  
\BibtexOrBiblatex
\electronicVersion
\PrintedOrElectronic

\ifpdf \usepackage[pdftex]{graphicx} \pdfcompresslevel=9
\else \usepackage[dvips]{graphicx} \fi

\usepackage{egweblnk} 
\usepackage{booktabs}  
\usepackage{mathptmx}
\usepackage{subcaption}
\usepackage{eucal}
\usepackage{amsmath}
\usepackage{amssymb}
\usepackage{xcolor}
\usepackage{bm}
\title[Makeup Extraction of 3D Representation via Illumination-Aware Image Decomposition]%
{Makeup Extraction of 3D Representation\\via Illumination-Aware Image Decomposition}

\author[Yang et al.]
{\parbox{\textwidth}{
\centering Xingchao Yang,$^{1,2}$\orcid{0000-0003-4736-1666}
         Takafumi Taketomi,$^{1}$\orcid{0000-0002-5353-0895} and
         Yoshihiro Kanamori$^{2}$\orcid{0000-0003-2843-1729}
        }
        \\
{\parbox{\textwidth}{\centering $^1$CyberAgent, AI Lab, Japan
         \quad$^2$University of Tsukuba, Japan
       }
}
}

\begin{document}

\teaser{
 \includegraphics[width=0.94\linewidth]{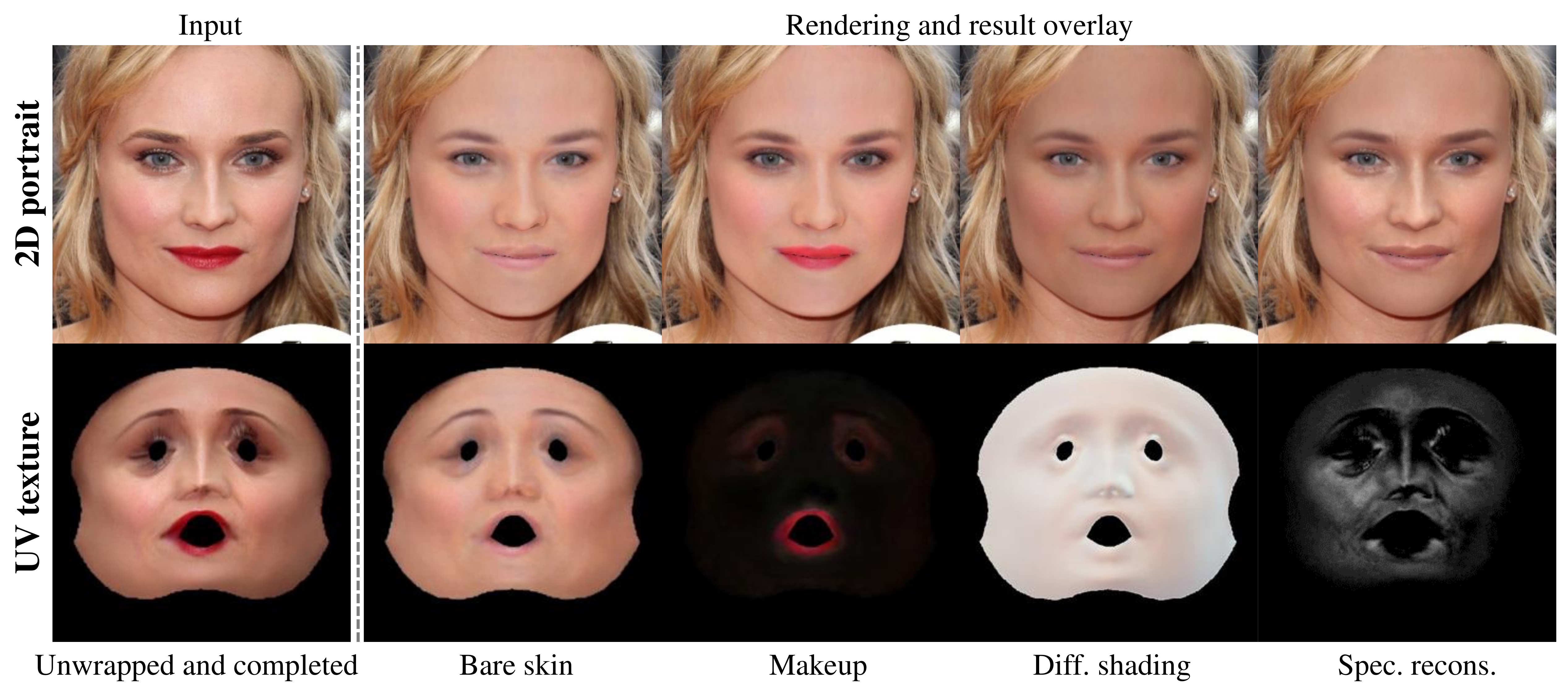}
 \centering
  \caption{
  Makeup-aware facial inverse rendering and component-wise reconstruction.
  The top row displays a makeup portrait input and overlaid rendering images (from left to right: bare skin only, bare skin plus makeup, bare skin multiplied by diffuse shading, and plus specular reconstruction) whereas the bottom row shows disentangled materials in the UV space.
  }
\label{fig:teaser}
}

\maketitle
\begin{abstract}
Facial makeup enriches the beauty of not only real humans but also virtual characters; therefore, makeup for 3D facial models is highly in demand in productions. However, painting directly on 3D faces and capturing real-world makeup are costly, and extracting makeup from 2D images often struggles with shading effects and occlusions. This paper presents the first method for extracting makeup for 3D facial models from a single makeup portrait. Our method consists of the following three steps. First, we exploit the strong prior of 3D morphable models via regression-based inverse rendering to extract coarse materials such as geometry and diffuse/specular albedos that are represented in the UV space.
Second, we refine the coarse materials, which may have missing pixels due to occlusions.
We apply inpainting and optimization.
Finally, we extract the bare skin, makeup, and an alpha matte from the diffuse albedo. 
Our method offers various applications for not only 3D facial models but also 2D portrait images. 
The extracted makeup is well-aligned in the UV space, from which we build a large-scale makeup dataset and a parametric makeup model for 3D faces. 
Our disentangled materials also yield robust makeup transfer and illumination-aware makeup interpolation/removal without a reference image.

\ccsdesc[500]{Computing methodologies~Computer graphics}
\ccsdesc[500]{Computing methodologies~Computer vision}
\ccsdesc[500]{Computing methodologies~Machine learning}

\printccsdesc   
\end{abstract}  
~\begin{figure*}[ht]
    \centering
    \includegraphics[width=\linewidth]{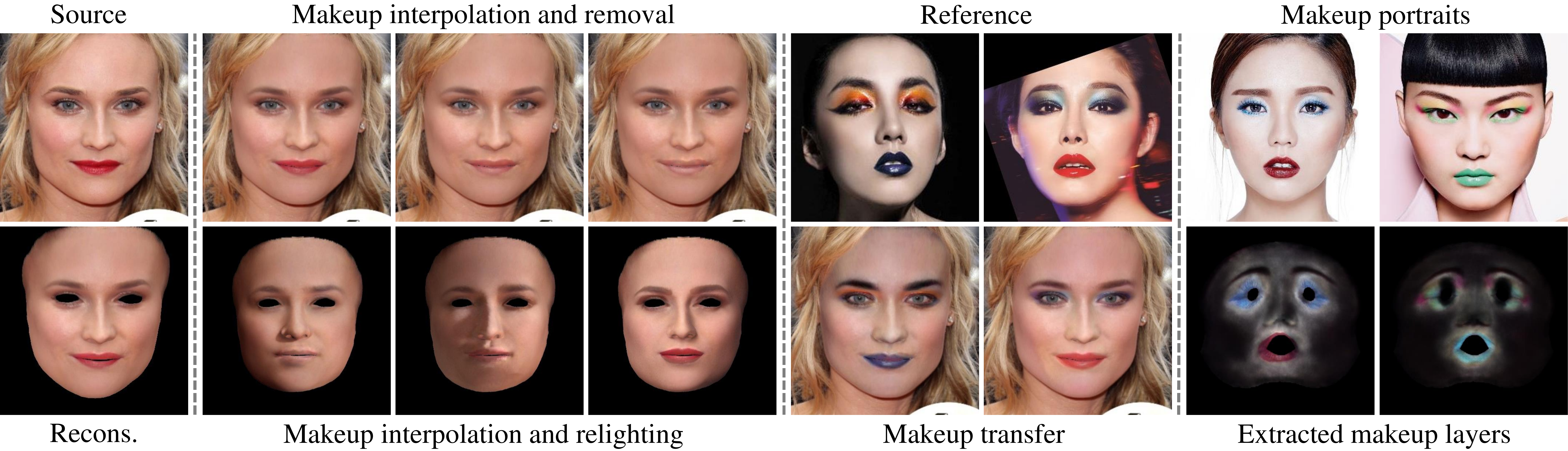}
    \caption{Example applications of our method. From left to right, makeup controllable 3D face reconstruction, 
    illumination-aware makeup interpolation/removal/relighting,
    illumination-aware makeup transfer, and makeup data collection.} 
\label{fig:introduction_app}
\end{figure*}

\section{Introduction}

Facial makeup is an art of enhancing human appearance, dating back to ancient times. Currently, it is quite commonly used for beautification purposes to improve the quality of life. 
Furthermore, facial makeup enriches the user experience of face-related applications such as VR/AR, video games, online commerce, and social camera apps.
These trends have been driving active research on facial makeup in computer graphics and computer vision. In particular, research on makeup for 3D facial models has been gaining increasing attention in the movie and advertising industries for digital humans.

To obtain facial makeup for 3D characters, the following three approaches are currently available; 1) direct painting, 2) capturing of real-world makeup, and 3) makeup transfer from 2D images. 
Direct painting is labor-intensive for makeup artists and quite costly.
Capturing real-world makeup requires special devices~\cite{Scherbaum11Makeup, PBCR} and thus is also costly and not scalable.
2D makeup transfer~\cite{Li:2018:MM, makeuptransferSurvey} is the current mainstream of facial makeup research, exploiting a myriad of facial makeup photos available on the Internet. 
However, most existing studies have focused on 2D-to-2D transfer and struggled with physical constraints. 
For example, faces of in-the-wild photos frequently contain lighting effects such as specular highlights and shadows and occlusions with hands, possibly with various facial expressions and head poses.
Recent 3D face reconstruction techniques~\cite{sfsnetSengupta18, egger20203d} can handle these physical constraints, but none of the existing techniques focus on facial makeup.

In this paper, we propose the first integrated solution for extracting makeup for 3D facial models from a single portrait image.
Fig.~\ref{fig:teaser} shows example outputs of our method.
Our method exploits the strong facial prior of the 3D morphable model (3DMM)~\cite{FLAME:SiggraphAsia2017} via regression-based inverse rendering and extracts coarse facial materials such as geometry and diffuse/specular albedos as UV textures.
Unlike the existing regression-based techniques for facial inverse rendering~\cite{sfsnetSengupta18, egger20203d}, we further refine the coarse facial materials via optimization for higher fidelity.
To alleviate the inherent skin color bias in the 3DMM, we also integrate skin color adjustment inspired by color transfer.
From the refined diffuse albedo, we extract the bare skin, facial makeup, and an alpha matte.
The alpha matte plays a key role in various applications such as manual tweaking of the makeup intensity and makeup interpolation/removal. 
The extracted makeup is well aligned in the UV space, from which we build a large-scale makeup texture dataset and a parametric makeup model using principal component analysis (PCA) for 3D faces.
By overlaying rendered 3D faces onto portrait images, we can achieve novel applications such as illumination-aware (\textit{i.e., relightable}) makeup transfer, interpolation, and removal, working on 2D faces (see Fig.~\ref{fig:introduction_app}).

The key contributions are summarized as follows:

\begin{itemize}
\item We present the first method to achieve illumination-aware makeup extraction for 3D face models from in-the-wild face images.
\item We propose a novel framework that improves each of the following steps; 
(1) an extended 3D face reconstruction network that infers not only diffuse shading but also specular shading via regression,
(2) a carefully designed inverse rendering method to generate high-fidelity textures without being restricted by the limited lighting setup, and
(3) a novel procedure that is specially designed for extracting makeup by leveraging the makeup transfer technique.
The UV texture representation effectively integrates these three modules into a single framework.
\item Our extracted illumination-independent makeup of the UV texture representation facilitates many makeup-related applications. The disentangled maps are also editable forms. We employ the extracted makeup to build a PCA-based makeup model that is useful for 3D face reconstruction of makeup portraits.
\end{itemize}

~\section{Related Work}
\label{sec:related_work}
Our framework consists of three steps for extracting makeup from a portrait. It is related to recent approaches in terms of three aspects. 
First, we discuss facial makeup-related research. Subsequently, we review intrinsic image decomposition which can separate the input image into several elements. Finally, we discuss 3D face reconstruction methods that generate a 3D face model from a single portrait image.

\subsection{Facial Makeup}
Facial makeup recommendation and makeup transfer methods have been developed in the computer graphics and computer vision research fields.
Makeup recommendation methods can provide appropriate makeup for faces. The method in \cite{Scherbaum11Makeup} captured 56 female faces with and without facial makeup. Suitable makeup was recommended by analyzing the principal components of the makeup and combining them with the facial appearance.
An examples-rules guided deep neural network for makeup recommendation has been designed~\cite{10.5555/3298239.3298377}. Professional makeup knowledge was combined with the corresponding before and after makeup data to train the network.
However, it is challenging to collect a large-scale makeup dataset by following the existing methods. 
Recently, we proposed BareSkinNet~\cite{bareskinnet}, which can remove makeup and lighting effects from input face images. However, BareSkinNet cannot be used for subsequent makeup applications because it discards the makeup patterns and specular reflection.
In contrast, our method can automatically extract the makeup layer information from portrait image inputs. As a result, a large-scale makeup dataset can be constructed.

The aim of makeup transfer is to transfer the makeup of a person to others. Deep learning technologies have significantly accelerated makeup transfer research. BeautyGAN ~\cite{Li:2018:MM} is a method based on CycleGAN~\cite{cyclegan} that does not require a before and after makeup image pair. A makeup loss function was also designed to calculate the difference between makeups. The makeup loss is a histogram matching loss that approximates the color distribution of the relative areas of different faces. Subsequent approaches that have used BeautyGAN as the baseline can solve more challenging problems. LADN~\cite{gu2019ladn} achieves heavy makeup transfer and removal, whereas PSGAN~\cite{jiang2019psgan} and SCGAN~\cite{Deng_2021_CVPR} solve the problem of makeup transfer under different facial expressions and head poses. CPM~\cite{m_Nguyen-etal-CVPR21} is a color and pattern transfer method and SOGAN~\cite{SOGAN} is a shadow and occlusion robust GAN for makeup transfer. EleGANt~\cite{yang2022elegant} is a locally controllable makeup transfer method. Makeup transfer can also be incorporated into the Virtual-Try-On applications~\cite{ca-gan, 10.1111:cgf.14456}. However, existing methods do not consider illumination and can only handle 2D images. Our method takes advantage of the makeup transfer technique. The extracted makeup is disentangled into bare skin, facial makeup, and illumination in the UV space. Furthermore, the extracted makeup is editable.

\subsection{Intrinsic Image Decomposition}
We mainly review the recent intrinsic image decomposition methods relating to the portrait image input. 
An intrinsic image decomposition method~\cite{SimulatingCVPR2015} has been employed, thereby enabling accurate and realistic makeup simulation from a photo.
SfSNet~\cite{sfsnetSengupta18} is an end-to-end network that decomposes face images into the shape, reflectance, and illuminance. This method uses real and synthetic images to train the network. Inspired by SfSNet, Relighting Humans~\cite{relighting_humans} attempts to infer a light transport map to solve the problem of light occlusion from an input portrait. The aforementioned methods can handle only diffuse reflection using spherical harmonic (SH)~\cite{sh} lighting. Certain methods use a light stage to obtain a large amount of portrait data with illumination~\cite{10.1145/3306346.3323008, Pandey:2021, 10.1145/3414685.3417824}. The global illumination can be inferred by learning these data. As data collection using a light stage requires substantial resources, several more affordable methods have been proposed~\cite{tajimaPG21, Lagunas2021SingleimageFH, ji2022relight, wimbauer2022rendering, Tan2022VoLuxGANAG, yeh2022learning}.
In this study, we focus on makeup with the aim of decomposing a makeup portrait into bare skin, makeup, diffuse, and specular layers without using light stage data.

\subsection{Image-Based 3D Face Reconstruction}
3D face reconstruction from a single-view portrait is challenging  because in-the-wild photos always contain invisible areas or complex illumination. Existing 3D face reconstruction methods~\cite{1227983, tran2016regressing, tewari2017mofa, tewari2017self, GuoTPAMI2018, genova2018unsupervised, RingNet:CVPR:2019, deng2020accurate, Shang2020Self, PracticalFaceReconsCGF, 10.1145/3450626.3459936, EMOCA, MICA} generally use a parametric face model (known as a 3DMM)~\cite{Blanz3DMM99, FLAME:SiggraphAsia2017, bfm, 10.1007/s11263-017-1009-7} to overcome this problem. In general, 3D face reconstruction is achieved by fitting the projected 3DMM to the input image. These methods estimate the SH lighting while inferring the shape and texture. We recommend that readers refer to ~\cite{egger20203d, 3dreconssurvey}, as these surveys provide a more comprehensive description of the 3DMM and 3D face reconstruction.

The 3DMM-based 3D face reconstruction method estimates coarse facial materials and cannot achieve a high-fidelity facial appearance.
A detailed 3D face reconstruction method was proposed in~\cite{PracticalFaceReconsCGF}, which can reconstruct the roughness and specular compared to the previous methods. This method involves the setup of a virtual light stage of illumination and the use of a two-stage coarse-to-fine technique to refine the facial materials. 
Inspired by ~\cite{PracticalFaceReconsCGF}, we employ coarse facial materials and take specular into consideration. Furthermore, we extend and train a deep learning network using the method in ~\cite{deng2020accurate} as the backbone with a large-scale dataset. We optimize the textures in UV space to refine facial materials, to solve the problem of self-occlusion and obstacles of the face. The completed UV texture is advantageous because the complete makeup can be extracted to achieve high-fidelity makeup reconstruction.

The generation of a completed facial UV texture is a technique for obtaining refined facial materials. Several methods use UV texture datasets to train an image-translation network for the generation of the completed texture via supervised learning~\cite{saito2017photorealistic, Yamaguchi:2018:ToG, deng2017uvgan, Gecer_2019_CVPR, 9156897, 9156897, Lattas_2020_CVPR, Bao:2021:ToG}. Other methods~\cite{Chen_Han_Shan_2022,Gecer_2021_CVPR} use the GAN inversion~\cite{gan_inversion} technique to generate faces in different directions for completion. However, both of these methods are limited by the training dataset. We adopt a state-of-the-art UV completion method known as DSD-GAN~\cite{dsd-gan} to obtain a completed high-fidelity face texture, which is to be used as the objective for the optimization of the refined facial materials. This method employs self-supervised learning to fill the missing areas without the need for paired training data.

~\section{Approach}
\label{sec:approach}

\begin{figure*}[t]
    \centering
    \includegraphics[width=\linewidth]{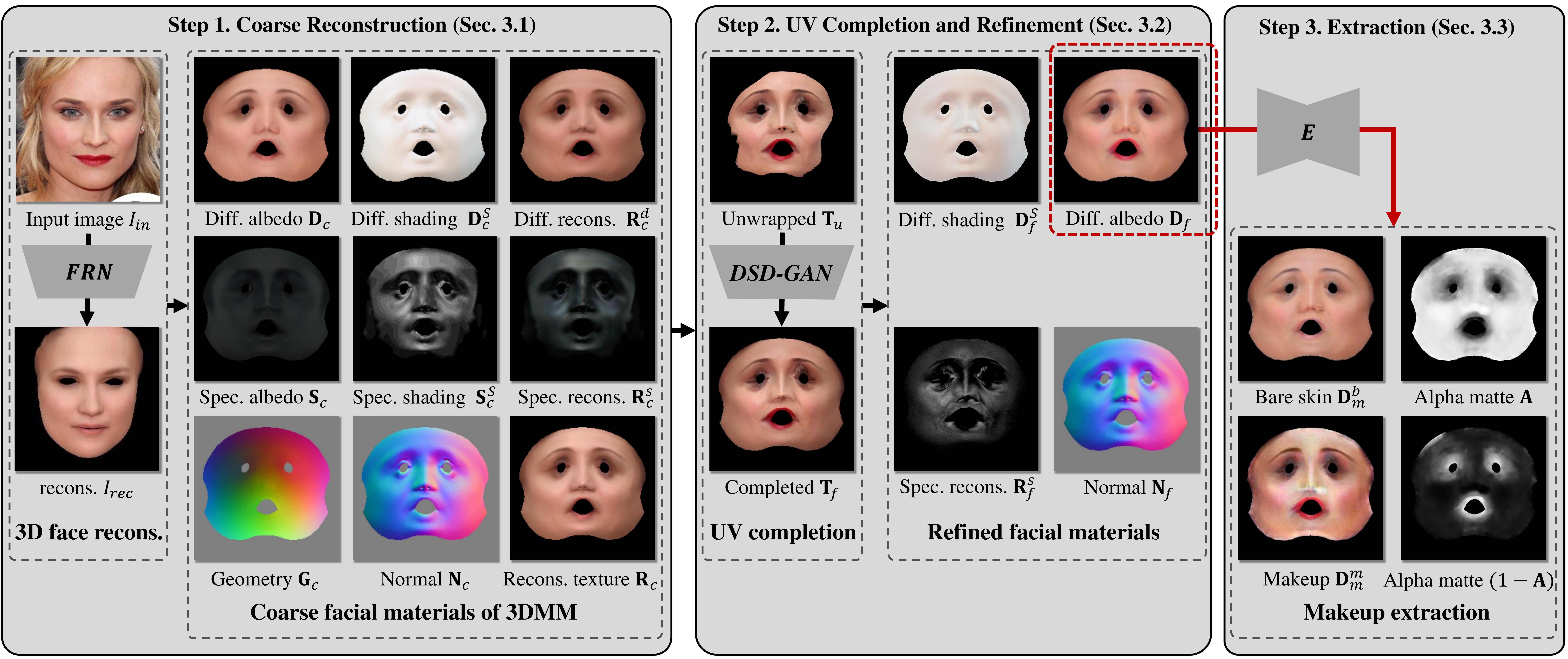}
    \caption{Overview of our framework.
    Given a makeup portrait, we extract an illumination-independent bare skin and makeup in the UV space via the following three steps;
    First, we reconstruct a 3D face to estimate the coarse facial materials using a 3D face reconstruction network $FRN$~\cite{deng2020accurate} (Sec.~\ref{sec:module_1}).
    Second, we refine the coarse facial materials, which may have missing pixels owing to occlusion. We apply an inpainting network DSD-GAN~\cite{dsd-gan} and then apply optimization (Sec.~\ref{sec:module_2}).
    Finally, we extract a bare skin, makeup, and an alpha matte from the refined diffuse albedo using makeup extraction network $E$ (Sec.~\ref{sec:module_3}).
    }
\label{fig:method_0}
\end{figure*}

We design a coarse-to-fine texture decomposition process. As shown in Fig.~\ref{fig:method_0}, our framework is composed of three steps.
1) We estimate the coarse 3D facial materials using 3D face reconstruction by 3DMM fitting. In order to handle highlights, we extend a general 3DMM fitting algorithm. The reconstructed coarse facial materials are used for the refinement process in the next step (Sec.~\ref{sec:module_1}).
2) We propose an inverse rendering method to obtain refined 3D facial materials via optimization. First, we use the 3DMM shape that is obtained in the previous step to sample the image in UV space. Subsequently, a UV completion method is employed to obtain a high-fidelity entire face texture. Finally, using the completed texture as the objective, and coarse facial materials as priors, the refined facial materials are optimized. (Sec.~\ref{sec:module_2}).
3) The refined diffuse albedo that is obtained from the previous step is used as the input. Inspired by the makeup transfer technique, we design a network to disentangle bare skin and makeup. The key idea is that the makeup albedo can be extracted using an alpha blending manner for bare skin and makeup (Sec.~\ref{sec:module_3}).
The details of each step are described in the following subsections.

\begin{figure*}[t]
    \centering
    \includegraphics[width=\linewidth]{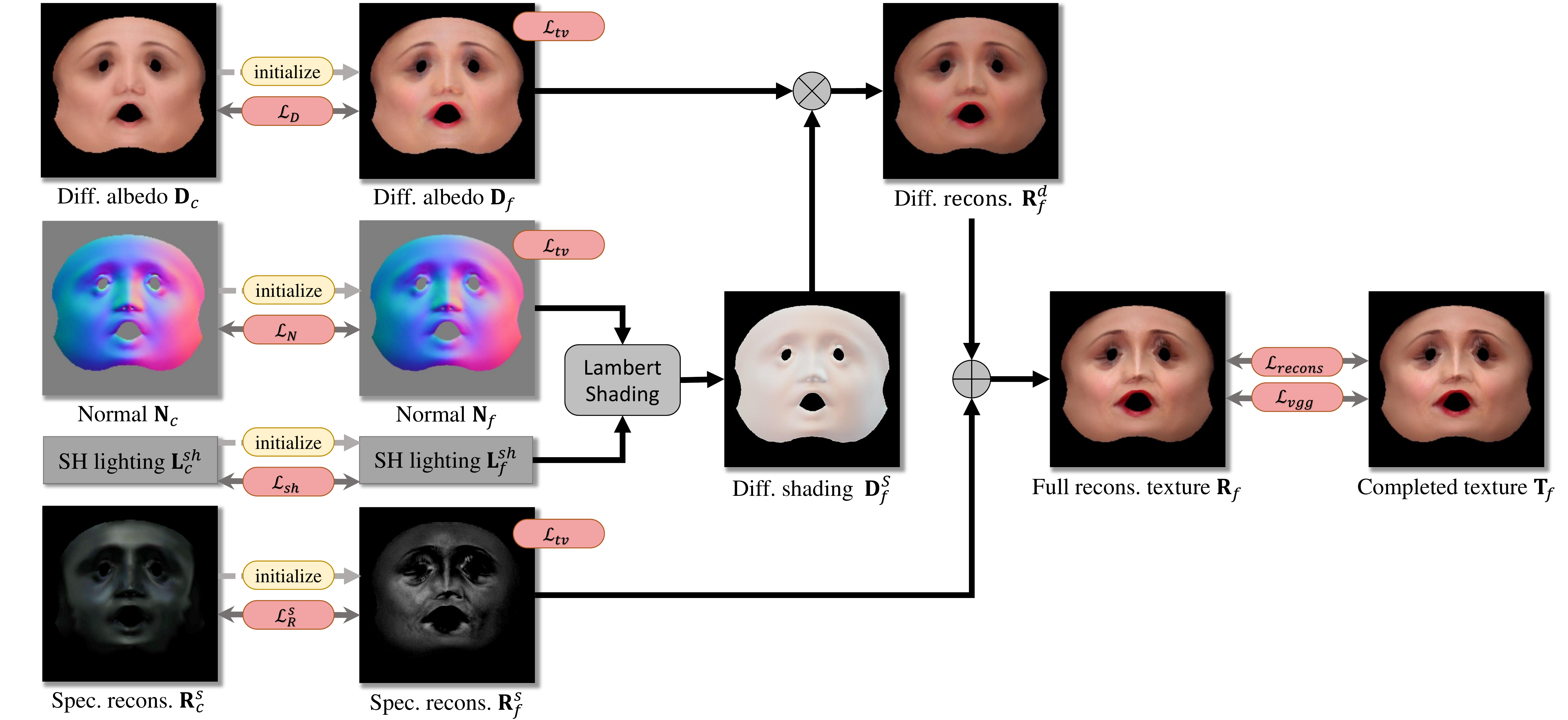}
    \caption{
    Facial material optimization module for Step 2 (Sec.~\ref{sec:module_2}).
    We optimize the coarse facial materials ${\mathbf{D}_c, \mathbf{N}_c, \mathbf{R}^s_c}$ and SH lighting $\mathbf{L}^{sh}_c$ so that the full reconstruction $\mathbf{R}_f$ resembles the completed texture $\mathbf{T}_f$.
    The refined diffuse albedo $\mathbf{D}_{f}$, normal $\mathbf{N}_{f}$, and specular reconstruction $\mathbf{R}_{f}^{s}$ are the outputs.
    $\oplus$ and $\otimes$ denote the per-pixel addition and multiplication, respectively.
    } 
\label{fig:method_2}
\end{figure*}


\subsection{Coarse Facial Material Reconstruction}
\label{sec:module_1}

We obtain the coarse facial materials using a 3D face reconstruction method based on regression-based inverse rendering.
We use the FLAME~\cite{FLAME:SiggraphAsia2017} model with specular albedo from AlbedoMM~\cite{Smith_2020_CVPR}. The diffuse and specular albedo of FLAME are defined in the UV texture space. We only use the facial skin region in our study. Compared to existing methods, we extend the capability of the 3D face reconstruction network~\cite{deng2020accurate} ($FRN$) to estimate the shape, diffuse albedo, and diffuse shading, as well as the specular albedo and specular shading. Inspired by~\cite{PracticalFaceReconsCGF},
a simplified virtual light stage with regular icosahedral parallel light sources is set up to infer the specular shading.
The intensity of 20 light sources is predicted during the reconstruction process, and the direction of the light sources can be adjusted slightly. Furthermore, In order to eliminate the limited color range of the skin tone of FLAME, we estimate the skin tone adjustment parameters to ensure a diffuse albedo that is similar to the original image. The skin tone ablation study is depicted in Fig.~\ref{fig:ablation_skin}. This process can improve the diversity of the diffuse albedo representation capability of FLAME.

The coarse shape geometry $\mathbf{G}_{c}$, diffuse albedo $\mathbf{D}_{c}$, and specular albedo $\mathbf{S}_{c}$ of the 3DMM are defined as follows:
\begin{align}
&\mathbf{G}_{c} = \bar{\mathbf{G}} + \mathbf{B}_{id}\mathbf{\alpha} + \mathbf{B}_{ex}\mathbf{\beta} ,\\
&\mathbf{D}_{c} = \bar{\mathbf{D}} + \mathbf{B}_{d}\mathbf{\gamma} \, \odot \, \mathbf{C}_{gain} + \mathbf{C}_{bias} , \\
&\mathbf{S}_{c} = \bar{\mathbf{S}} + \mathbf{B}_{s}\mathbf{\delta} ,
\end{align}
where $\bar{\mathbf{G}}$, $\bar{\mathbf{D}}$, and $\bar{\mathbf{S}}$ are the average geometry, diffuse albedo, and specular albedo, respectively. ${\odot}$ denotes the Hadamard product. The subscript $c$ indicates coarse facial materials. Moreover, $\mathbf{B}_{id}$, $\mathbf{B}_{ex}$, $\mathbf{B}_{d}$, and $\mathbf{B}_{s}$ are the PCA basis vectors of the identity, expression, diffuse albedo, and specular albedo, respectively, whereas $\mathbf{\alpha}\in\mathbb{R}^{200}$, $\mathbf{\beta}\in\mathbb{R}^{100}$, $\mathbf{\gamma}\in\mathbb{R}^{100}$, and $\mathbf{\delta}\in\mathbb{R}^{100}$ are the corresponding parameters for controlling the geometry and reflectance of a 3D face. Finally, $\mathbf{C}_{gain}$ and $\mathbf{C}_{bias}$ are the skin tone adjustment parameters. 

Our reconstructed texture can be formulated as follows:
\begin{align}
\mathbf{R}_{c} &= \mathbf{R}_{c}^{d} + \mathbf{R}_{c}^{s} \\
& = \mathbf{D}_{c} \odot \, \mathbf{D}_{c}^{S} + \mathbf{S}_{c} \odot \, \mathbf{S}_{c}^{S} ,
\end{align}
where $\mathbf{R}_{c}^{d}$, $\mathbf{R}_{c}^{s}$, $\mathbf{D}_{c}^{S}$, and $\mathbf{S}_{c}^{S}$ are the diffuse reconstruction, specular reconstruction, diffuse shading, and specular shading, respectively.
In this paper, a geometrically-derived shading component is referred to as ``shading,'' wheras a multiplication with reflectance is dubbed ``reconstruction.''
Using the normal $\mathbf{N}_{c}$ and second-order SH lighting coefficients $\mathbf{L}_{c}^{sh}\in\mathbb{R}^{27}$, 
the diffuse shading is calculated following the Lambertian reflectance model~\cite{lambertian}. The normal $\mathbf{N}_{c}$ is computed from the geometry $\mathbf{G}_{c}$.
The specular shading is calculated following the Blinn–Phong reflection model~\cite{Blinn-Phong-Model}. We define the 20 light sources of the virtual light stage with the light intensity $\mathbf{L}_{i}\in\mathbb{R}^{20}$ and light direction $\mathbf{L}_{d}\in\mathbb{R}^{60}$. $\mathbf{\rho}\in\mathbb{R}^{20}$ is the exponent that controls the shininess.
We employ the 3D face reconstruction network implementation of ~\cite{deng2020accurate}, and extend it to regress the parameters $\mathbf{\chi} = (\mathbf{\alpha}, \mathbf{\beta}, \mathbf{\gamma}, \mathbf{\delta},   \mathbf{C}_{gain}, \mathbf{C}_{bias}, \mathbf{r}, \mathbf{t}, \mathbf{L}_{c}^{sh}, \mathbf{L}_{i}, \mathbf{L}_{d}, \mathbf{\rho})$, where $\mathbf{r}\in\mathbb{R}^{3}$ is the face rotation, and $\mathbf{t}\in\mathbb{R}^{3}$ is the face translation. Using the geometry $\mathbf{G}_c$, reconstructed texture $\mathbf{R}_c$, rotation $\mathbf{r}$, and translation $\mathbf{t}$, the reconstructed image ${I}_{rec}$ can be rendered.

We refer to~\cite{deng2020accurate} to set up the loss functions for the model training as follows:
\begin{equation}
\begin{aligned}
\mathcal{L}_{c}(\mathbf{\chi}) &= 
\omega_{photo} \, \mathcal{L}_{photo}(\mathbf{\chi}) + \omega_{lan} \, \mathcal{L}_{lan}(\mathbf{\chi}) \\
&+ \omega_{skin} \, \mathcal{L}_{skin}(\mathbf{C}_{gain}, \mathbf{C}_{bias}) \\
&+ \omega_{reg} \, \mathcal{L}_{reg}(\mathbf{\alpha}, \mathbf{\beta}, \mathbf{\gamma}, \mathbf{\delta}, \mathbf{L}_{i}, \mathbf{L}_{d}) ,
\end{aligned}
\end{equation}
where $\mathcal{L}_{photo}(\mathbf{\chi})$ is the L1 pixel loss between the skin region of input image ${I}_{in}$ and reconstructed image ${I}_{rec}$. $\mathcal{L}_{lan}(\mathbf{\chi})$ is the L2 loss between detected landmarks from ${I}_{in}$ and the projected landmarks of the 3DMM. $\mathcal{L}_{skin}(\mathbf{C}_{gain}, \mathbf{C}_{bias})$ is the L1 loss for computing the mean color error between the skin region of ${I}_{in}$ and diffuse albedo $\mathbf{D}_{c}$. This is our specially designed loss term to adjust the skin tone. $\mathcal{L}_{reg}(\mathbf{\alpha}, \mathbf{\beta}, \mathbf{\gamma}, \mathbf{\delta}, \mathbf{L}_{i}, \mathbf{L}_{d})$ is the regulation loss for preventing a failed face reconstruction result. In contrast to the previous method~\cite{deng2020accurate}, we extend constraints on the specular related coefficients of $\mathbf{\delta}$, $\mathbf{L}_i$, and $\mathbf{L}_d$. $\mathcal{L}_{reg}(\mathbf{\alpha}, \mathbf{\beta}, \mathbf{\gamma}, \mathbf{\delta}, \mathbf{L}_{i}, \mathbf{L}_{d})$ is determined as follows:
\begin{equation}
\begin{aligned}
\mathcal{L}_{reg} &= 
\omega_{\alpha}\Vert{\mathbf{\alpha}}\Vert^{2}_{2}
+\omega_{\beta}\Vert{\mathbf{\beta}}\Vert^{2}_{2}
+\omega_{\gamma}\Vert{\mathbf{\gamma}}\Vert^{2}_{2} \\
&+\omega_{\delta}\Vert{\mathbf{\delta}}\Vert^{2}_{2}
+\omega_{L}\Vert{\mathbf{L}_i}\Vert^{2}_{2}
+ \omega_{L}\Vert{\mathbf{L}_d}\Vert^{2}_{2} ,
\end{aligned}
\end{equation}
where $\Vert\cdot\Vert_{2}$ denotes the L2 norm. The balance weights are set to 
$\omega_{photo}=19.2$, 
$\omega_{lan}=5$, 
$\omega_{skin}=3$, 
$\omega_{reg}=3 \times 10^{-4}$, 
$\omega_{\alpha}=1.0$,
$\omega_{\beta}=0.8$,
$\omega_{\gamma}=1.7 \times 10^{-2}$,
$\omega_{\delta}=1.0$, and $\omega_{L}=1.0$ in all experiments.


\begin{figure*}[t]
    \centering
    \includegraphics[width=\linewidth]{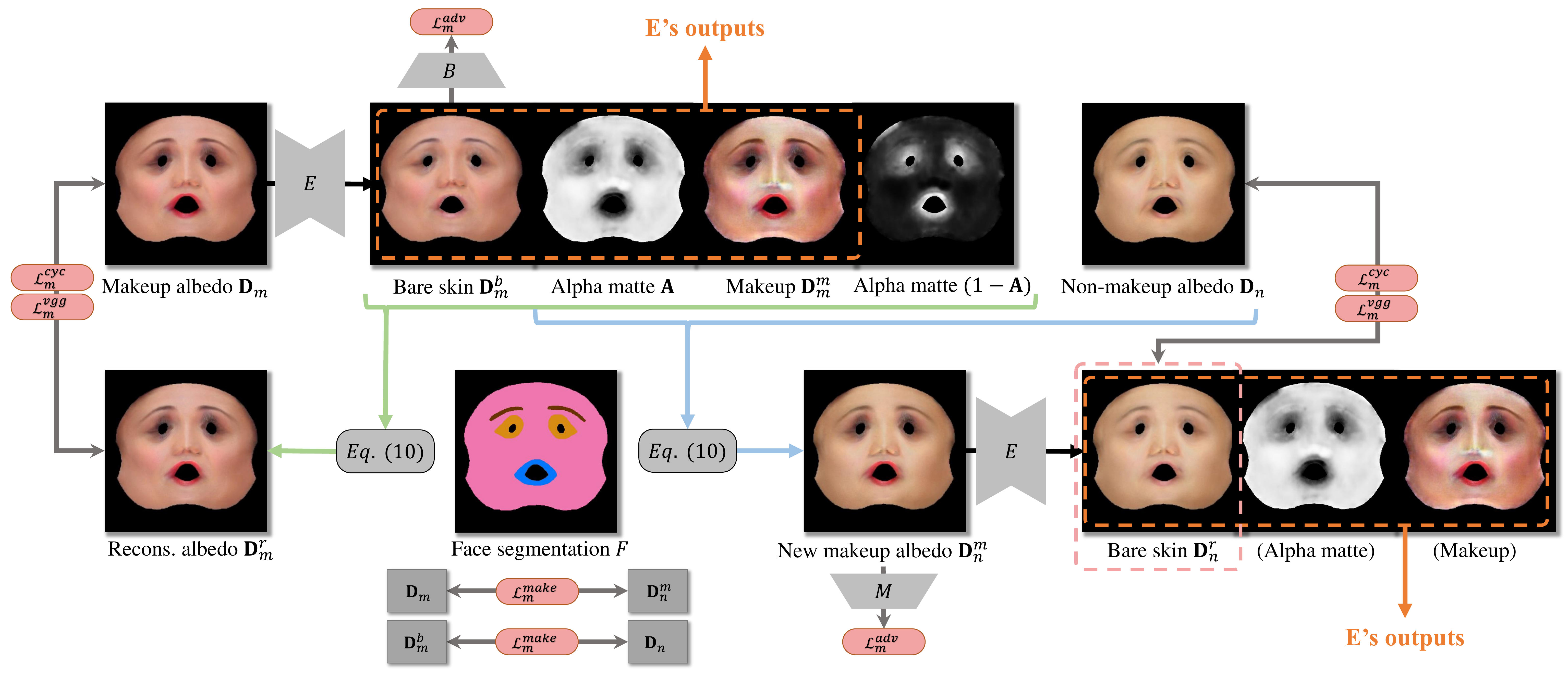}
    \caption{Makeup extraction with network $E$ for Step 3 (Sec.~\ref{sec:module_3}).
    Considering alpha blending in mind, the network $E$ decomposes the refined diffuse albedo into bare skin, alpha matte, and makeup.
    } 
\label{fig:method_3}
\end{figure*}

\subsection{UV Completion and Facial Material Refinement}
\label{sec:module_2}

The goal of this step is to obtain the disentangled refined facial materials.
We use the geometry $\mathbf{G}_{c}$ to sample the colors from the input image ${I}_{in}$ and project them to the UV texture space $\mathbf{T}_{u}$. $\mathbf{T}_{u}$ contains the missing area owing to self-occlusion or obstacles. As opposed to the 3D vertex color sampling approach used in related work~\cite{m_Nguyen-etal-CVPR21, Chen_Han_Shan_2022}, we adopt the image-to-UV rendering approach used in DSD-GAN~\cite{dsd-gan} to obtain a high-quality texture. The direct use of incomplete textures will cause many problems, such as noise and error. Thereafter, we fill the missing areas following DSD-GAN to obtain the completed UV texture $\mathbf{T}_{f}$. The subscript $f$ indicates refined facial materials. The difference between our method and DSD-GAN is that we use the FLAME model for implementation, whereas DSD-GAN uses the BFM model~\cite{bfm}. The results of the UV completion are presented in Fig.~\ref{fig:result_process}.

Using the completed UV texture as an objective with facial details, the optimization-based refinement module (see Fig.~\ref{fig:method_2}) is designed. Given the target texture $\mathbf{T}_{f}$, the coarse prior $\mathbf{D}_{c}$, $\mathbf{N}_{c}$, $\mathbf{L}_{c}^{sh}$, and $\mathbf{R}_{c}^{s}$ are used for initialization, and we optimize the refined materials of $\mathbf{D}_{f}$, $\mathbf{N}_{f}$, $\mathbf{L}_{f}^{sh}$, and $\mathbf{R}_{f}^{s}$ to reconstruct the refined texture $\mathbf{R}_{f}$. The diffuse shading is calculated following the Lambertian reflectance model~\cite{lambertian}. For the specular, we directly optimize the specular reconstruction $\mathbf{R}_{f}^{s}$, rather than the specular albedo and specular shading, which is not restricted by the light source settings of the virtual light stage in Sec.~\ref{sec:module_1}. Note that the specular reconstruction $\mathbf{R}_{c}^{s}$ is three channels image due to the specular albedo $\mathbf{S}_{c}$ of 3DMM, while $\mathbf{R}_{f}^{s}$ is converted to one channel image for stable and efficient optimization.

The loss functions for the optimization are calculated as follows:
\begin{align}
\mathcal{L}_{f}(\Psi) &=
\omega_{recons}\mathcal{L}_{recons}(\mathbf{R}_{f})
+ \omega_{vgg}\mathcal{L}_{vgg}(\mathbf{R}_{f}) \notag \\
&+ \omega_{tv}\mathcal{L}_{tv}(\mathbf{D}_{f}, \mathbf{N}_{f}, \mathbf{R}_{f}^{s}) \\
&+ \omega_{prior}\mathcal{L}_{prior}(\mathbf{D}_{f}, \mathbf{N}_{f}, \mathbf{R}_{f}^{s}, \mathbf{L}_{f}^{sh}) , \notag
\end{align}
where $\Psi = (\mathbf{R}_{f}, \mathbf{D}_{f}, \mathbf{N}_{f}, \mathbf{R}_{f}^{s}, \mathbf{L}_{f}^{sh})$ is a new parameter set for optimization. $\mathcal{L}_{recons}(\mathbf{R}_{f})$ ensures L1 consistency between $\mathbf{R}_{f}$ and $\mathbf{T}_{f}$. $\mathcal{L}_{vgg}(\mathbf{R}_{f})$ is the perceptual loss~\cite{vggloss} that aims to preserve the facial details. $\mathcal{L}_{tv}(\mathbf{D}_{f}, \mathbf{N}_{f}, \mathbf{R}_{f}^{s})$ is the total variation loss~\cite{tvloss} that encourages spatial smoothness in the optimized textures $\mathbf{D}_{f}$, $\mathbf{N}_{f}$, and $\mathbf{R}_{f}^{s}$.
\begin{equation}
\begin{aligned}
\mathcal{L}_{prior}(\mathbf{D}_{f}, \mathbf{N}_{f}, \mathbf{R}_{f}^{s}, \mathbf{L}_{f}^{sh}) &=
\omega_{D} \, \mathcal{L}_{D}(\mathbf{D}_{f}) + \omega_{N} \, \mathcal{L}_{N}(\mathbf{N}_{f}) \\
&+ \omega_{R}^{s} \, \mathcal{L}_{R}^{s}(\mathbf{R}_{f}^{s}) + \omega_{sh} \, \mathcal{L}_{sh}(\mathbf{L}_{f}^{sh}) ,
\end{aligned}
\end{equation}
where $\mathcal{L}_{prior}(\mathbf{D}_{f}, \mathbf{N}_{f}, \mathbf{R}_{f}^{s}, \mathbf{L}_{f}^{sh})$ regulates the optimized texture to be similar to the coarse prior. We compute the L1 loss over the diffuse albedo, normal, and specular reconstruction as $\mathcal{L}_{D}(\mathbf{D}_{f})$, $\mathcal{L}_{N}(\mathbf{N}_{f})$, and $\mathcal{L}_{R}^{s}(\mathbf{R}_{f}^{s})$, respectively. The coarse textures are resized to the same resolution as that of refined textures. We note that the coarse and refined textures are not exactly equal; they differ in clarity, and with or without makeup. Therefore, before calculating the losses between the coarse and refined textures, we simply use a Gaussian blur filter $K$ to blur the refined textures in each iteration. We use the blurred refined texture to calculate the loss, while the original refined texture is not changed. $\mathcal{L}_{sh}(\mathbf{L}_{f}^{sh})$ is the L2 loss over the SH lighting. We normalize $\mathbf{N}_{f}$ to [-1, 1] following each optimization iteration to guarantee the correctness of the normal. The parameters 
$\omega_{recons}=40$, 
$\omega_{vgg}=5$,
$\omega_{tv}=10$,
$\omega_{prior}=1.0$,
$\omega_{D}=4$,
$\omega_{N}=1.0$,
$\omega_{R}^{s}=1.0$,
and $\omega_{sh}=1.0$
balance the importance of the terms. The refined textures of our coarse-to-fine optimization step are illustrated in Figs.~\ref{fig:result_process}, ~\ref{fig:result_final}, and ~\ref{fig:result_albedo}.


\subsection{Makeup Extraction}
\label{sec:module_3}

In this step, 
we only use the refined diffuse albedo $\mathbf{D}_{f}$, and decompose the texture into the makeup, bare skin, and an alpha matte. To train the network, we create two diffuse albedo datasets with and without makeup following the previous process; the makeup albedo $\mathbf{D}_{m}$ and non-makeup albedo $\mathbf{D}_{n}$, respectively.

As shown in Fig.~\ref{fig:method_3}, we design a makeup extraction network based on alpha blending. Our network consists of a makeup extractor ${E}$, a makeup discriminator ${M}$, and a bare skin discriminator ${B}$. ${M}$ and ${B}$ attempt to distinguish the makeup and non-makeup image. The core idea is that we regard the diffuse albedo as a combination of makeup and bare skin that is achieved by alpha blending. Therefore, ${E}$ extracts the bare skin $\mathbf{D}_{m}^{b}$, makeup $\mathbf{D}_{m}^{m}$, and alpha matte $\mathbf{A}$. 
$\mathbf{A}$ is used to blend the extracted makeup $\mathbf{D}_{m}^{m}$ and a non-makeup albedo $\mathbf{D}_{n}$ to generate a new makeup albedo $\mathbf{D}_{n}^{m}$ which has the identity from $\mathbf{D}_{n}$ and makeup from $\mathbf{D}_{m}$. 
The reconstructed makeup albedo $\mathbf{D}_{m}^{r}$ and the reconstructed bare skin $\mathbf{D}_{n}^{r}$ should be consistent with their original input.
The reconstructed $\mathbf{D}_{m}^{r}$ and generated $\mathbf{D}_{n}^{m}$ are formulated as follows:
\begin{equation}
\begin{aligned}
\mathbf{D}_{m}^{r} = \mathbf{A} \odot \mathbf{D}_{m}^{b} + (1 - \mathbf{A}) \odot \mathbf{D}_{m}^{m} ,\\
\mathbf{D}_{n}^{m} = \mathbf{A} \odot \mathbf{D}_{n} + (1 - \mathbf{A}) \odot \mathbf{D}_{m}^{m} , \label{eq:alhpa_blend}
\end{aligned}
\end{equation}
where $(1 - \mathbf{A})$ is an inverted version of $\mathbf{A}$ with pixel values of $[0,1]$.
The discriminators ensure that the generated bare skin $\mathbf{D}_{m}^{b}$ and makeup $\mathbf{D}_{n}^{m}$ are reliable. 
As the network takes advantage of the uniformity of the UV space, the makeup transfer is straightforward, and the problem of face misalignment does not need to be considered. 

The loss function is given by:
\begin{equation}
\begin{aligned}
\mathcal{L}_{m}(\Phi) &= 
\omega_{m}^{cyc}\mathcal{L}_{m}^{cyc}(\mathbf{D}_{m}^{r}, \mathbf{D}_{n}^{r})
+ \omega_{m}^{vgg}\mathcal{L}_{m}^{vgg}(\mathbf{D}_{m}^{r}, \mathbf{D}_{n}^{r}) \\
&+ \omega_{m}^{adv}\mathcal{L}_{m}^{adv}(\mathbf{D}_{m}^{b}, \mathbf{D}_{n}^{m})
+ \omega_{m}^{tv}\mathcal{L}_{m}^{tv}(\mathbf{D}_{m}^{b}, \mathbf{D}_{n}^{m}) \\
&+ \omega_{m}^{make}\mathcal{L}_{m}^{make}(\mathbf{D}_{m}^{b}, \mathbf{D}_{n}^{m}) ,
\end{aligned}
\end{equation}
where $\Phi = (\mathbf{D}_{m}^{r}, \mathbf{D}_{n}^{r}, \mathbf{D}_{m}^{b}, \mathbf{D}_{n}^{m})$, $\mathcal{L}_{m}^{cyc}(\mathbf{D}_{m}^{r}, \mathbf{D}_{n}^{r})$ and $\mathcal{L}_{m}^{vgg}(\mathbf{D}_{m}^{r}, \mathbf{D}_{n}^{r})$ are the L1 loss and perceptual loss for the reconstruction, respectively. 
$\mathcal{L}_{m}^{adv}(\mathbf{D}_{m}^{b}, \mathbf{D}_{n}^{m})$ is the adversarial loss for the discriminators and generators. 
$\mathcal{L}_{m}^{tv}(\mathbf{D}_{m}^{b}, \mathbf{D}_{n}^{m})$ is the total variation loss for $\mathbf{D}_{m}^{b}$ and $\mathbf{D}_{m}^{m}$ to provide smooth texture generation, whereas $\mathcal{L}_{m}^{make}(\mathbf{D}_{m}^{b}, \mathbf{D}_{n}^{m})$ is the makeup loss introduced by BeautyGAN~\cite{Li:2018:MM}. 
We defined the corresponding face region ${F}$ of the brows, eyes, and lips in the UV texture space to compute makeup loss, and the compared textures are $\mathbf{D}_{m}$ and $\mathbf{D}_{n}^{m}$, and $\mathbf{D}_{n}$ and $\mathbf{D}_{m}^{b}$. 
Note that, in contrast with previous makeup transfer methods, the skin region is not calculated for the makeup loss because we believe that the difference in skin tone between individuals cannot be considered as makeup, while it can also be a restriction if the foundation of the makeup changes the entire face color. 
This is a trade-off between skin tone and makeup color. In this work, we assume that the makeup will not drastically change the skin tone. Moreover, as indicated in Fig.~\ref{fig:result_final}, even without the makeup loss of the skin region, our network precisely extracts the makeup of the cheeks using the alpha blending approach because we use two discriminators to distinguish the makeup and non-makeup textures. The details are discussed in Sec.~\ref{sec:experiments}.

We use $\omega_{m}^{cyc}=20$, $\omega_{m}^{vgg}=2$, $\omega_{m}^{adv}=5$, $\omega_{m}^{tv}=8$, and $\omega_{m}^{make}=1$ as the balancing terms. Our makeup extraction results are presented in Figs.~\ref{fig:result_process}, ~\ref{fig:result_final}, and ~\ref{fig:result_albedo}.
~\section{Implementation Details}
\label{sec:implementation_detail}

\subsection{Coarse Facial Material Reconstruction} 

We followed the training strategy of ~\cite{deng2020accurate} and use the same datasets for approximately 260K face images. We used~\cite{facealignment} to detect and crop the faces for alignment. The image size was $256 \times 256$ and the resolution of FLAME albedo textures was $256 \times 256$. We initialized the network with the weights of the pre-trained~\cite{ILSVRC15} and modified the last layer to estimate our own 3DMM parameters. The batch size was 8, and the learning rate was $1 \times 10^{-4}$ using an Adam optimizer with 20 training epochs. For the shininess parameter $\mathbf{\rho}$, we set the initial value to 200 to achieve highlight effects.

\subsection{UV Completion and Facial Material Refinement}

We sampled approximately 100K images from FFHQ~\cite{FFHQ} and CelebA-HQ~\cite{karras2018progressive} in the UV space to train DSD-GAN for the UV completion, and the resolution of the UV textures was $512 \times 512$. Prior to sampling, the face images were segmented using the method of ~\cite{Yu_2018_ECCV} and only the skin region of the face was used. Subsequently, we performed a manual cleanup to remove the low-quality textures.
Eventually, 60,073 textures remained for training.

The optimization-based refinement process was executed with 500 iterations for each texture. The learning rate of the Adam optimizer was set to $1 \times 10^{-2}$ with a 0.1 learning rate decay. The kernel size of the Gaussian blur filter $K$ was set to 11. 
 
\subsection{Makeup Extraction}

We combined two makeup datasets, namely the MT dataset~\cite{Li:2018:MM} and LADN dataset~\cite{gu2019ladn}, which consisted of 3,070 makeup images and 1,449 non-makeup images. A total of 300 makeup images were randomly selected for testing. By implementing the previous steps, both the makeup and non-makeup images were processed into albedo textures to train our network. We trained the network with 40 epochs and batch size of 1. The Adam optimizer used a learning rate of $1 \times 10^{-4}$. The makeup extractor ${E}$ had the same architecture as the generator of DSD-GAN, and PatchGAN ~\cite{isola2017image} was used for the discriminators $M$ and $B$.

We used Nvdiffrast~\cite{nvidiaffrast} for the differentiable renderer and trained the networks using a single NVIDIA GeForce RTX 2080 Ti GPU. Our training required approximately 3 days for the coarse facial reconstruction network and approximately 1 day for the makeup extraction network. Approximately 1 minute was required to process a texture in the refinement step.
~\section{Experiments}
\label{sec:experiments}

\begin{figure*}[t]
    \centering
    \includegraphics[width=0.95 \linewidth]{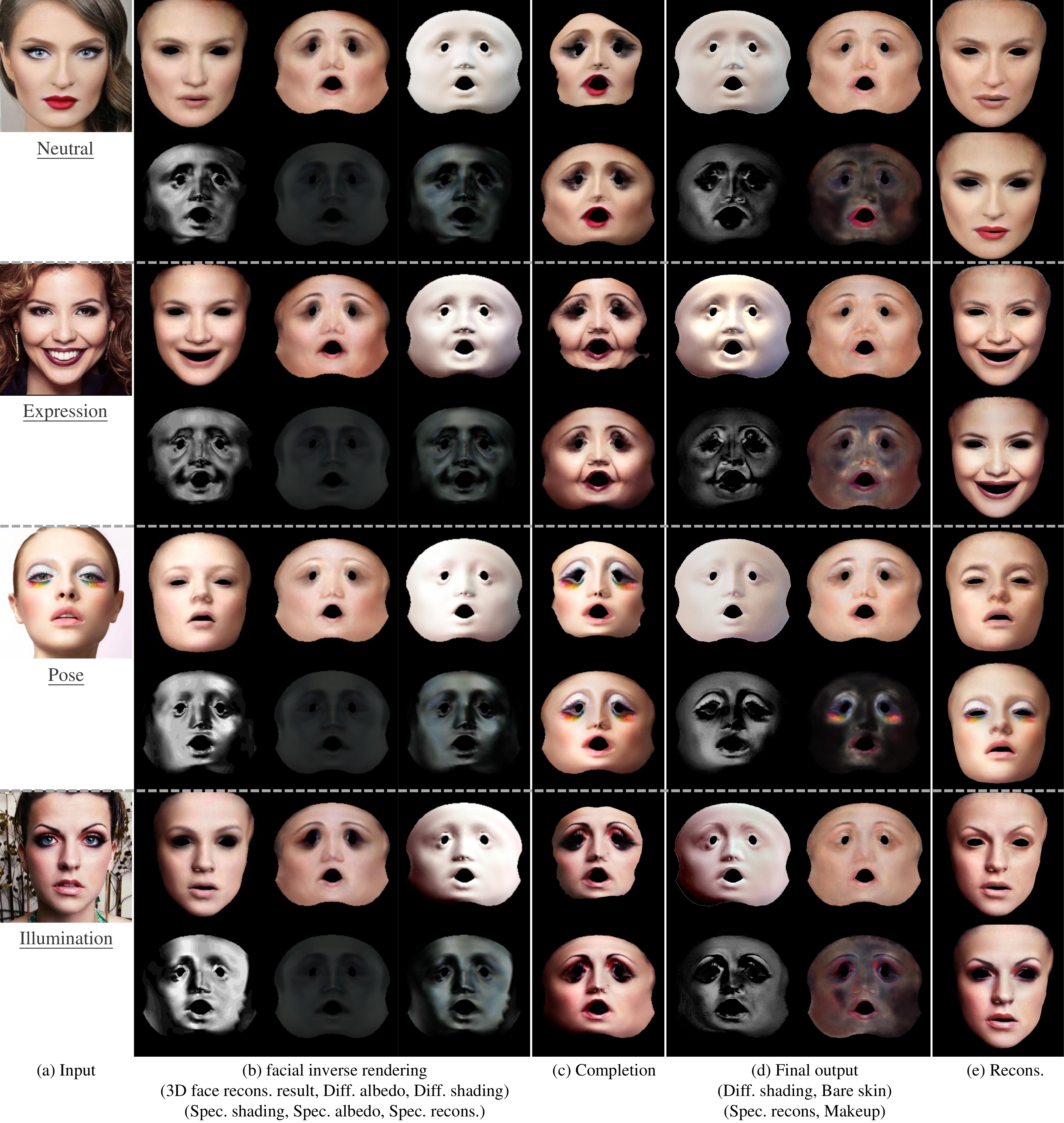}
    \caption{Intermediate outputs for each step. Left to right and top to bottom: (a) input makeup portraits,
    (b) facial inverse rendering (diffuse albedo, diffuse shading, specular shading, specular albedo, and specular reconstruction),
    (c) unwrapped and completed textures, 
    (d) final refined facial materials (diffuse shading, bare skin, specular reconstruction, and makeup),
    and (e) rendered bare skin face and rendered makeup face using textures from (d).} 
\label{fig:result_process}
\end{figure*}

\begin{figure}[t]
    \centering
    \includegraphics[width=\linewidth]{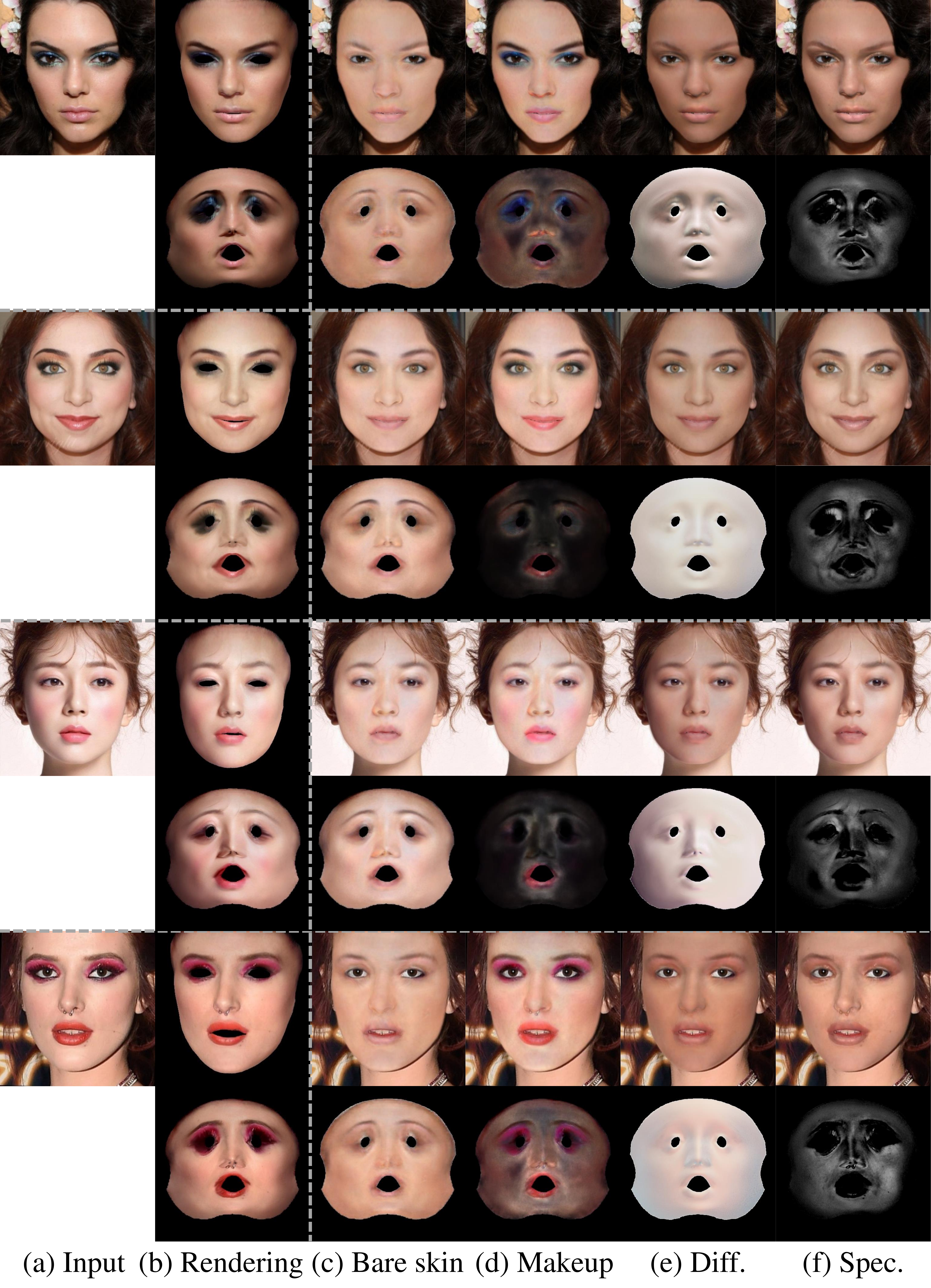}
    \caption{
    Final outputs of our framework.
    (a) Input makeup portraits, and (b) fully reconstructed renderings and corresponding textures.
    The columns from (c) to (f) are similar to those of Fig.~\ref{fig:teaser}.
    } 
\label{fig:result_final}
\end{figure}

\begin{figure}[t]
    \centering
    \includegraphics[width=\linewidth]{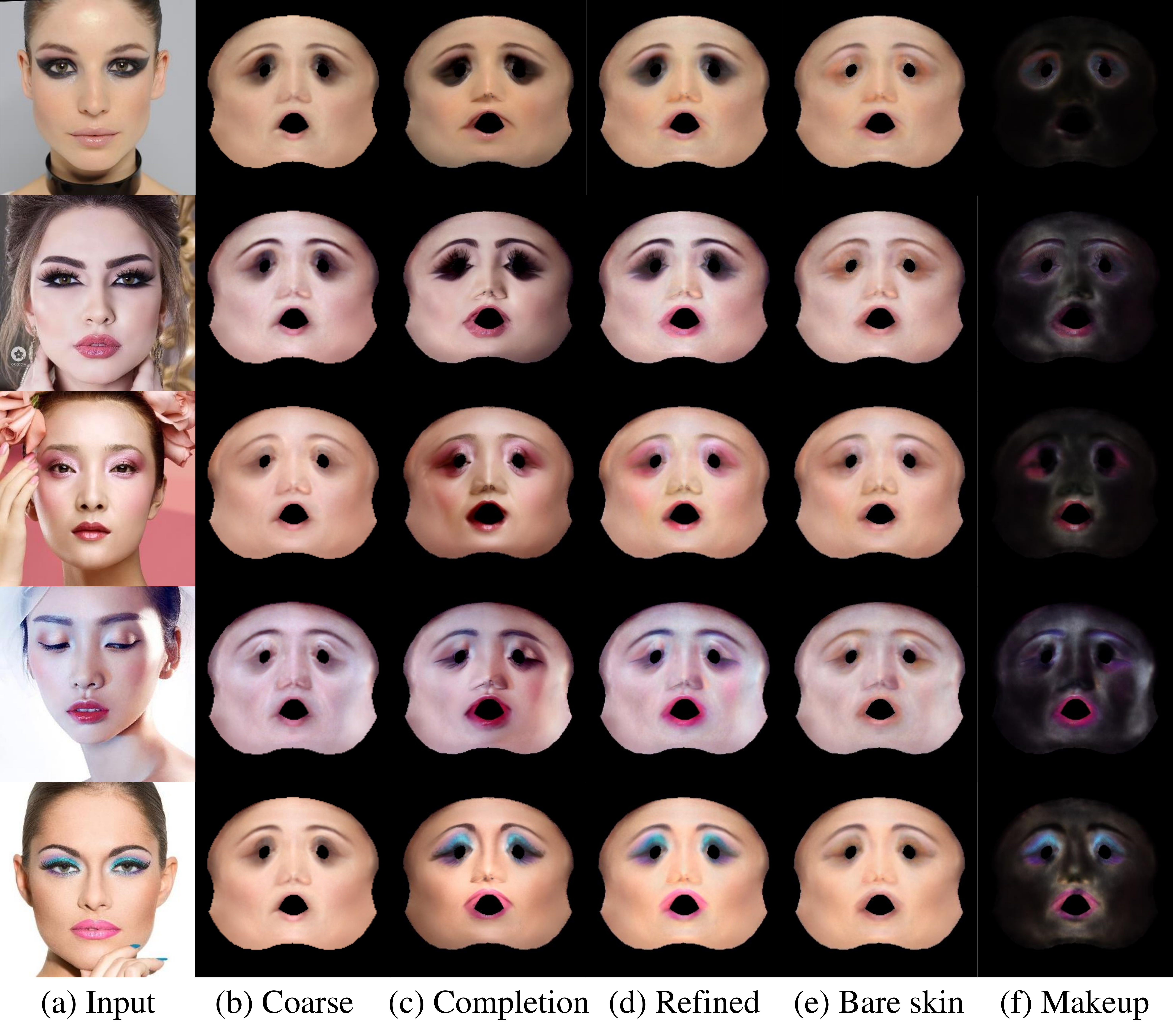}
    \caption{
    Texture outputs relating to makeup.
    Left to right: (a) input makeup portraits, (b) coarse diffuse albedo, (c) completed UV textures, (d) refined diffuse albedo, (e) bare skin, and (f) makeup.} 
\label{fig:result_albedo}
\end{figure}

\begin{figure}[t]
    \centering
    \includegraphics[width=\linewidth]{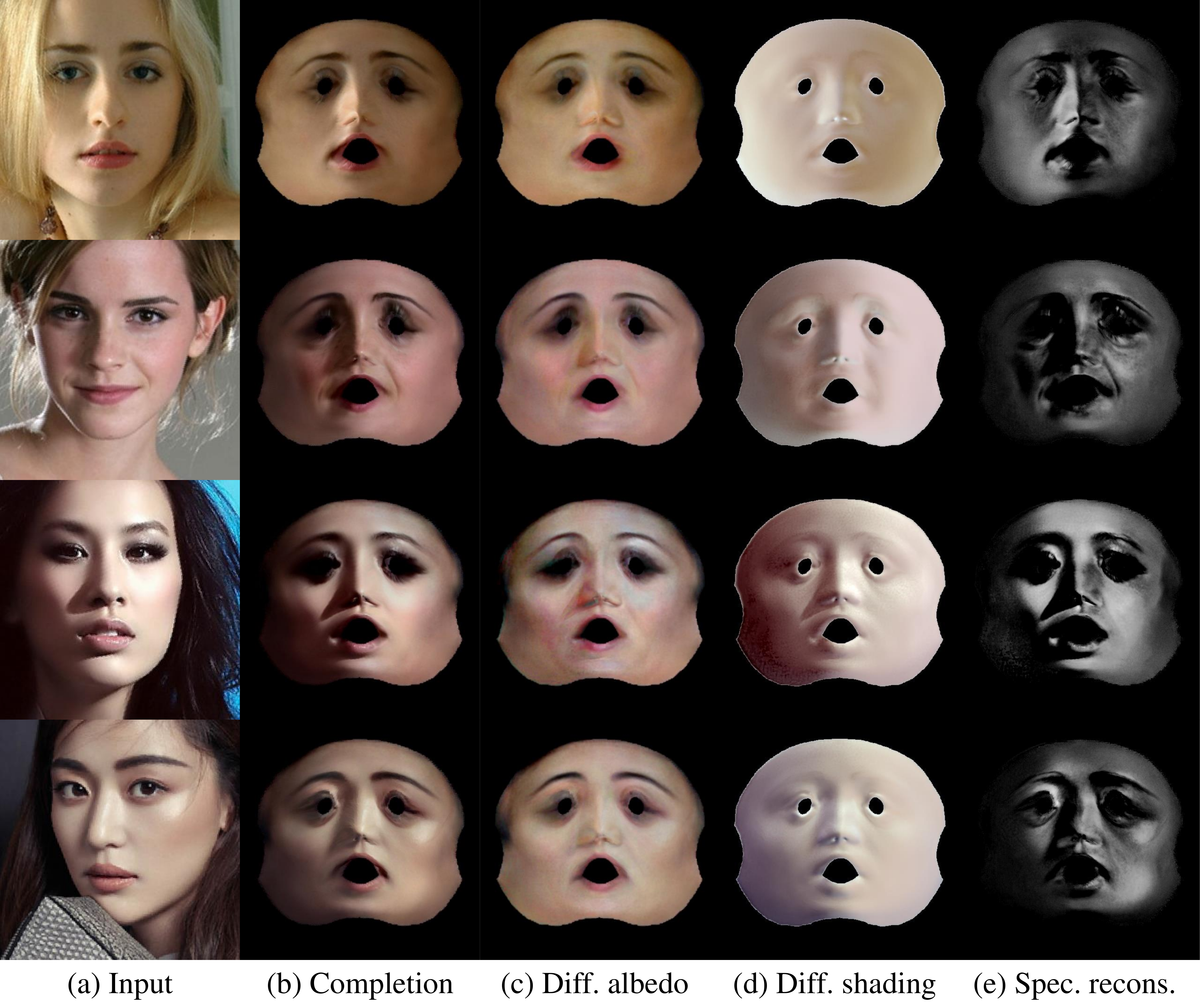}
    \caption{Outputs of decomposed refined facial materials in complex illumination conditions. Left to right: (a) input makeup portraits, (b) completed UV textures, (c) diffuse albedo, (d) diffuse shading, and (e) specular reconstruction.} 
\label{fig:result_illumination}
\end{figure}

 We evaluated the results of our approach. First, we present the intermediate outputs of each step of the framework (see Fig.~\ref{fig:result_process}). Thereafter, we discuss the final outputs (see Fig.~\ref{fig:result_final}). Subsequently, we analyze the albedo texture that was associated with the makeup and observe how the makeup changed in the albedo texture (see Fig.~\ref{fig:result_albedo}), Finally, we provide several examples with complex illumination and examine the decomposition (see Fig.~\ref{fig:result_illumination}). Note that the original images of the specular reconstruction were too dark to display, so we adjusted the contrast for better display. Please refer to the supplemental material to see the original images.

\noindent\textbf{Intermediate outputs of each step.} 
Fig.~\ref{fig:result_process} depicts the intermediate outputs, which are related to the final textures in (d), to illustrate the effectiveness of each step. Four makeup portraits are presented in (a) with different facial features: a neutral face, a face with expression, a face pose with an angle, and a face with uneven illumination. The results of the 3D face reconstruction are shown in (b). It can be observed that the diffuse albedo of the 3DMM contained only a coarse texture without makeup. The specular shading was affected by the limited light source, and thus, the global illumination could not be recovered. Furthermore, the specular albedo was a coarse texture; therefore, the final specular reconstruction is not detailed. Although it was not possible to obtain a reconstruction of the makeup using only the 3D face reconstruction method, these coarse facial materials were helpful for subsequent steps. 
We used the original texture and completed it to obtain the refined facial materials as well as makeup. The UV completion results are presented in (c). The final textures were refined using the outputs of the previous steps, as illustrated in (d). For better visualization, we show the blended makeup, which was calculated by $(1 - \mathbf{A}) \odot \mathbf{D}_{m}^{m}$. It can be observed that the makeup and bare skin were disentangled, the diffuse shading was adjusted, and the specular reconstruction was more detailed. 
We combined the final textures to achieve effects such as rendering a bare skin face and a makeup face while the lighting remained invariant. These results demonstrate that our method is robust to different makeup, expressions, poses, and illumination.

\noindent\textbf{The final outputs.} 
The final outputs are depicted in Fig.~\ref{fig:result_final}. We rendered the disentangled final UV textures separately for improved visualization: (c) bare skin, (d) bare skin with makeup, (e) bare skin with diffuse shading, and (f) bare skin with the full illumination model. Furthermore, we rendered the makeup faces in (b) using these textures which could be considered as a 3D face reconstruction of makeup portraits. It can be observed that the layers of bare skin, makeup, diffuse, and specular are disentangled. For example, a comparison of (c) and (d) indicates that makeup was added while no illumination was involved, and the makeup around the eyebrows, eyes, and lips was disentangled. The presence of makeup on the cheeks can also be clearly observed in the third identity from the top.

\noindent\textbf{Makeup in diffuse albedo.} 
The textures relating to the makeup are depicted in Fig.~\ref{fig:result_albedo} to demonstrate the effect of the makeup changes on the diffuse albedo textures. This process represents the main concept under consideration for extracting the makeup. The reconstructed coarse diffuse albedo could not preserve the makeup effectively; although the reconstruction results exhibited some black makeup around the eyes, it was difficult to preserve the makeup beyond the scope of the 3DMM texture space (see (b)). Thus, we used the original texture directly. The completed UV textures, which contained makeup and involved illumination, are depicted in (c). Subsequently, the illumination was removed. A comparison of (d), (e), and (f) demonstrates the validity of our makeup extraction network. 
Moreover, the final row presents an example of a makeup portrait with occlusion, for which our method was still effective in extracting the makeup. As we only used the sampled skin region while excluding the others, the missing regions were filled to become complete.

\noindent\textbf{Decomposition of illumination.} 
We evaluate the results of the refined facial materials to demonstrate the capabilities of our illumination-aware makeup extraction. It can be observed from the outputs in Fig.~\ref{fig:result_illumination} that portraits containing uneven lighting and shadows (particularly around the bridge of the nose) were captured by diffuse shading, whereas the highlights were reflected in the specular reconstruction. Thus, the diffuse albedo textures became clean and flat, and our makeup extraction was more precise after removing the illumination.

The entire process of our method is executed in the UV space. UV-represented makeup offers numerous advantages. First, 3D makeup avatar creation becomes accessible when the same UV coordinates are used. Furthermore, such makeup can be extended to scanned 3D faces, in which case makeup without illumination will be helpful. Second, the makeup can be further divided into several parts using a corresponding face segmentation mask, which will enable specification of which makeup region is to be used. Third, the textures can be directly edited and incorporated into a traditional rendering pipeline. Finally, the disentangled makeup maps can be collected to form a makeup dataset, which will be useful for 3D makeup face reconstruction or makeup recommendation. We explore several applications in Sec.~\ref{sec:application}.

~\section{Ablation Studies}
\label{sec:ablation}

\begin{figure}[t]
    \centering
    \includegraphics[width=\linewidth]{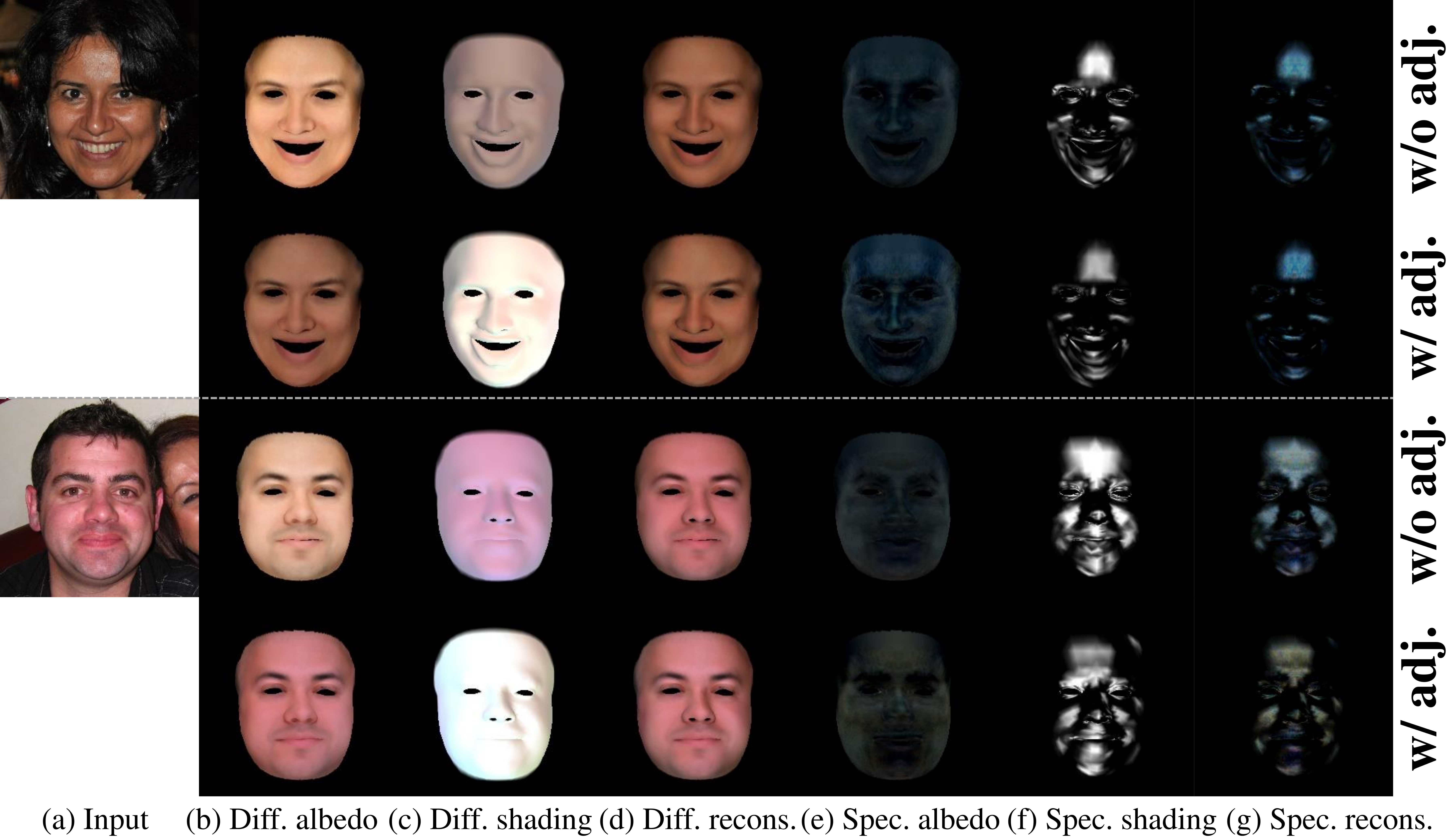}
    \caption{
    Qualitative ablation study for skin tone adjustment. For each identity, the upper and lower rows display the results before and after the skin tone adjustment, respectively.
    Left to right: (a) input face images, (b) diffuse albedo, (c) diffuse shading, (d) diffuse reconstruction, (e) specular albedo, (f) specular shading, and (g) specular reconstruction.
    } 
\label{fig:ablation_skin}
\end{figure}

\begin{table}[t]
\caption{
Quantitative ablation study for makeup face reconstruction results.
}
\label{tab:table_1}
\begin{center}
\begin{tabular}{l|ccc}
\toprule
 \ \ Complete data & RMSE & SSIM & LPIPS\\
\hline
\midrule
$FRN$ (w/o specular) & 0.056 & 0.953 & 0.091 \\
$FRN$ (w/ specular) & \ \ \ 0.053 $\downarrow$ & \ \ \ 0.956 $\uparrow$ & \ \ \ 0.088 $\downarrow$  \\
Rendered result & \color{gray}0.060 & \ \ \ \color{gray}0.968 $\Uparrow$ & \ \ \ \color{gray}0.062 $\Downarrow$ \\
\bottomrule
\toprule
 \ \ \ \ \ \ \ Test data & RMSE & SSIM & LPIPS\\
\hline
\midrule
$FRN$ (w/o specular) & 0.058 & 0.948 & 0.102 \\
$FRN$ (w/ specular) & \ \ \ 0.055 $\downarrow$ & \ \ \ 0.951 $\uparrow$ & \ \ \ 0.099 $\downarrow$  \\
Rendered result & 0.063 & \ \ \ 0.964 $\Uparrow$ & \ \ \ 0.066 $\Downarrow$ \\
\bottomrule
\end{tabular}
\setlength{\belowcaptionskip}{10pt}
\end{center}

\end{table}

We conducted ablation studies on the 3D face reconstruction step. We performed a comparative experiment with and without skin tone adjustment and assessed the influence on the components of 3DMM. We selected two images from FFHQ~\cite{FFHQ} with strong illumination. As illustrated in Fig.~\ref{fig:ablation_skin}, the skin tone adjustment had only a slight effect, or none at all, on the specular. Although the diffuse reconstruction did not change, the balance of the diffuse albedo and diffuse shading was significantly adjusted. Limited by the 3DMM texture, the diffuse albedos of the two faces before adjusting the skin tones had almost similar colors, resulting in an incorrect estimation of the diffuse shading. This would be misleading for the subsequent refinement and makeup extraction step. To mitigate this error, we adjusted the skin tone so that the average color was the same as that of the original image.

Tab.~\ref{tab:table_1} displays the results of the quantitative evaluation of the 3D makeup face reconstruction. We trained two face reconstruction networks with different illumination models: one used SH lighting without specular estimation, and the other was the full model. The rendered results using the bare skin, makeup, diffuse shading, and specular reconstruction textures are also listed for reference. 
We used the complete makeup, which was not used to train the 3D face reconstruction network for the general evaluation. We used a gray color to mark the rendered results, because the generated maps for rendering were trained on the complete dataset. Furthermore, the test makeup dataset, which was not used to train the makeup extraction network, was separated to evaluate the rendered results. The network containing the specular illumination model improved the accuracy of the reconstruction, thereby demonstrating the effectiveness of our model. Our final rendered makeup face exhibited improvement in terms of the perceptual similarity. Thus, the reconstructed results were more compatible with the visual evaluation, as we believe that makeup significantly influences human perception.

~\begin{figure}[t]
    \centering
    \includegraphics[width=\linewidth]{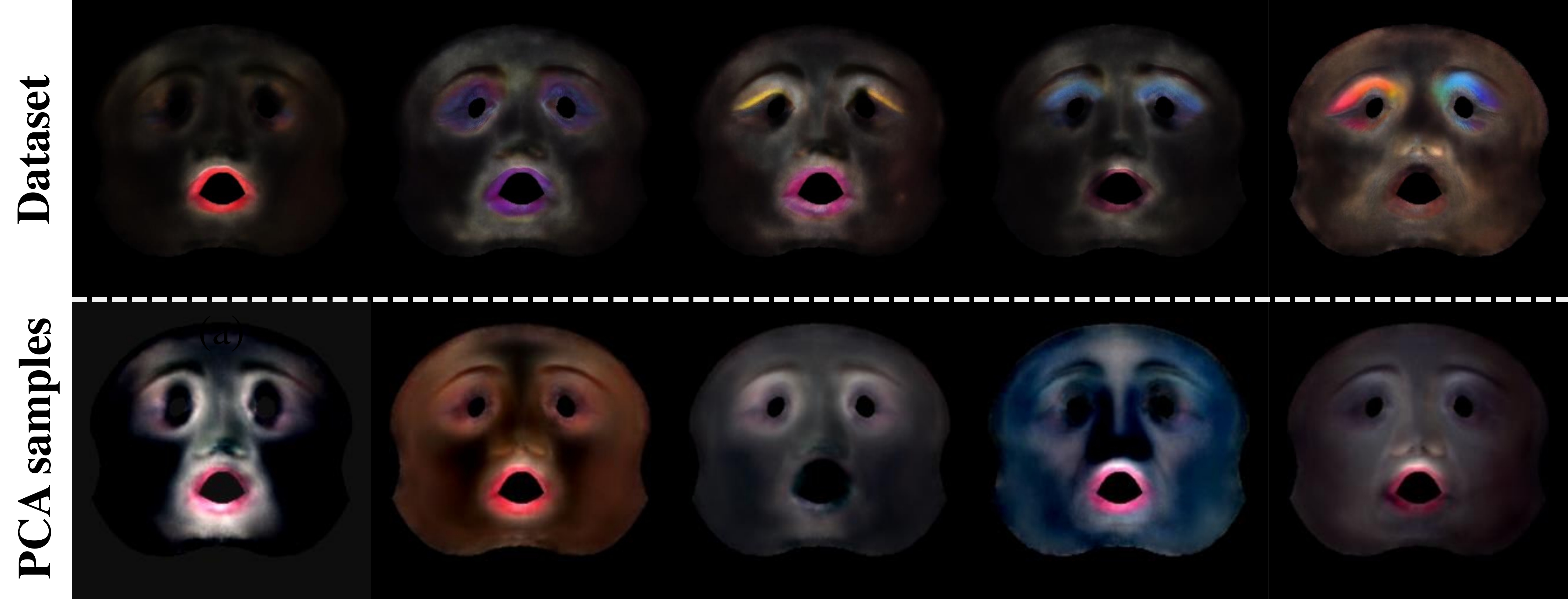}
    \caption{Makeup dataset and PCA-based makeup model. We used the dataset to create a PCA-based makeup model. The top row presents the extracted makeup textures from image input, and the bottom row shows randomly sampled makeup textures along the principal components from the makeup model.} 
\label{fig:result_pca_layer}
\end{figure}

\begin{figure*}[t]
    \centering
    \includegraphics[width=0.95\linewidth]{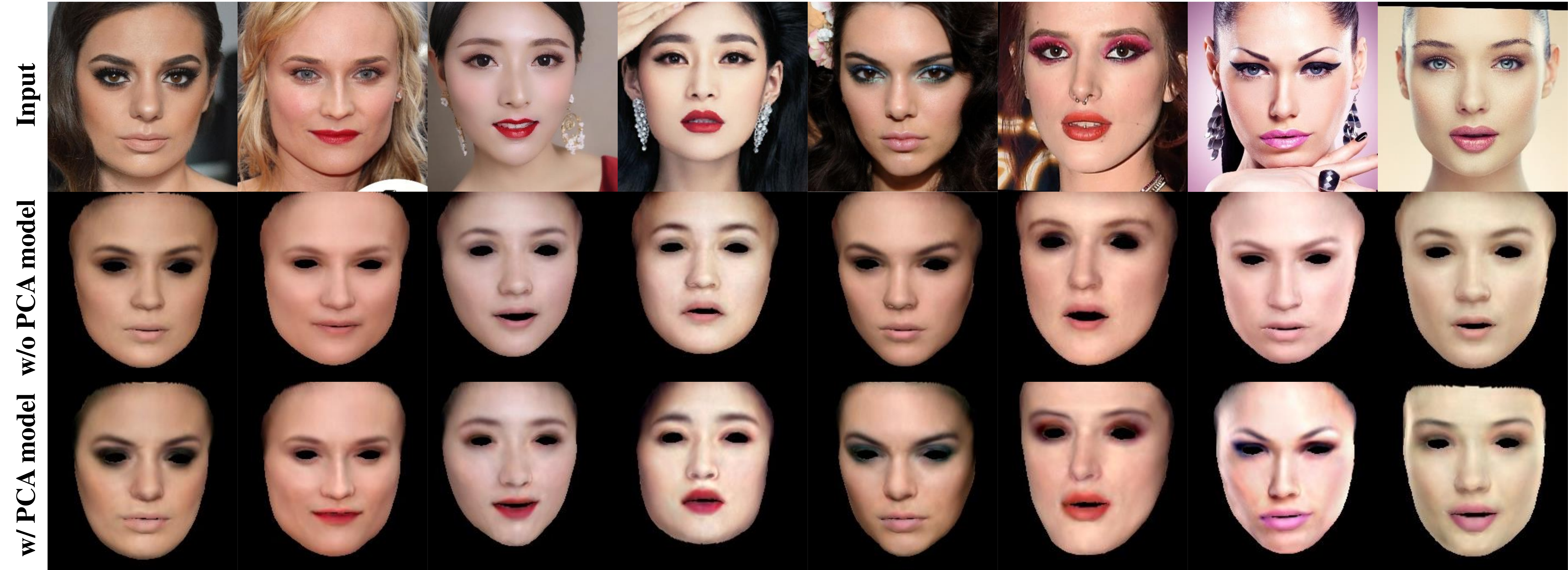}
    \caption{3D face reconstruction results using 3DMM.
    The PCA-based makeup model significantly improves the fidelity of synthesized makeup.
    } 
\label{fig:result_pca}
\end{figure*}

\begin{table*}[ht]
\caption{
Taxonomy of state-of-the-art makeup transfer methods.
“Misalignment”: faces with different poses. “Shade”: controlling the shade. “Control”: selection of the face region to be transferred. “Edit”: editing within arbitrary areas. "Occlusion": robust to occlusion. "Illumination": controlled illumination during transfer.
}
\label{tab:table_function}
\begin{center}
\begin{tabular}{l|cccccc}
\toprule
Method& Misalignment & Shade & Control & Edit & Occlusion & Illumination\\
\hline
\midrule
BeautyGAN~\cite{Li:2018:MM}& & & & & &\\
LADN~\cite{gu2019ladn}& &\checkmark & & & &\\
PSGAN~\cite{jiang2019psgan}&\checkmark &\checkmark &\checkmark & & &\\
SCGAN~\cite{Deng_2021_CVPR}&\checkmark &\checkmark & \checkmark& & &\\
CPM~\cite{m_Nguyen-etal-CVPR21}&\checkmark &\checkmark &\checkmark &\checkmark & &\\
SOGAN~\cite{SOGAN}&\checkmark &\checkmark &\checkmark &\checkmark &\checkmark &\\
EleGANt~\cite{yang2022elegant}&\checkmark &\checkmark &\checkmark &\checkmark & \checkmark &\\
Ours&\checkmark &\checkmark &\checkmark &\checkmark &\checkmark & \checkmark\\
\bottomrule
\end{tabular}
\setlength{\belowcaptionskip}{10 pt}
\end{center}
\end{table*}

\begin{figure*}[t]
    \centering
    \includegraphics[width=0.95\linewidth]{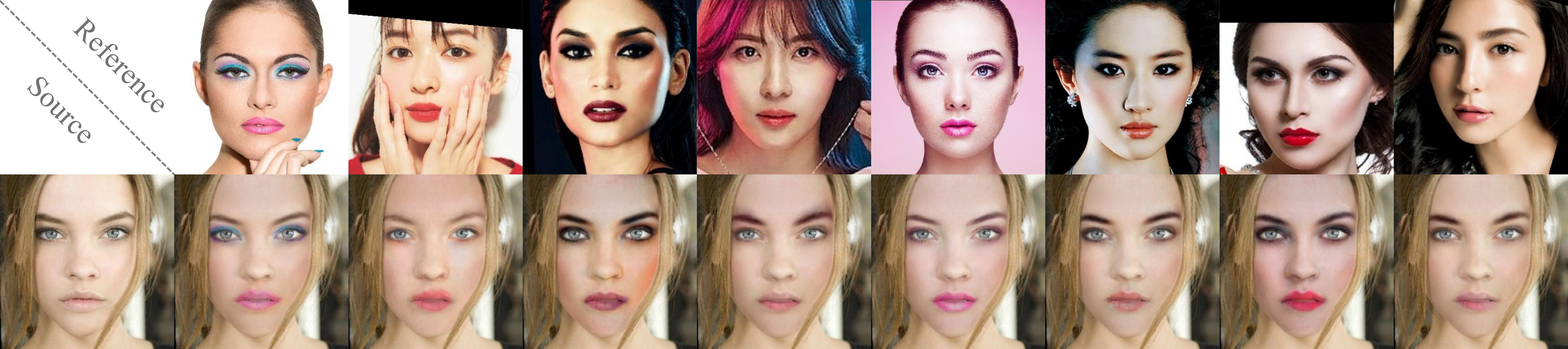}
    \caption{Our makeup transfer can handle occlusion, illumination, and face misalignment.} 
\label{fig:result_complex_transfer}
\end{figure*}

\begin{figure*}[t]
    \centering
    \includegraphics[width=0.95\linewidth]{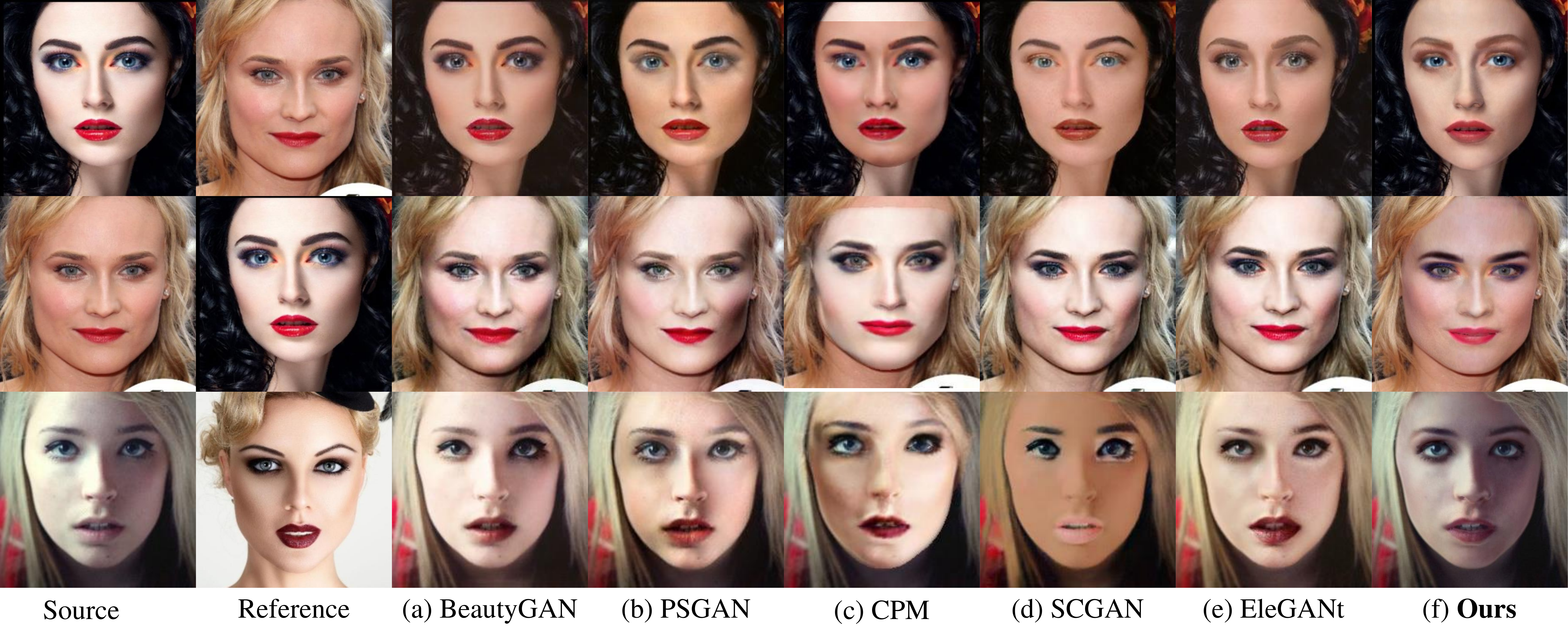}
    \caption{Qualitative comparison of makeup transfer between two faces. Left to right:
    source images providing the identity and illumination, the reference images providing the makeup,
    (a) BeautyGAN~\cite{Li:2018:MM}, (b) PSGAN~\cite{jiang2019psgan}, (c) CPM~\cite{m_Nguyen-etal-CVPR21}, (d) SCGAN~\cite{Deng_2021_CVPR}, (e) EleGANt~\cite{yang2022elegant}, and (f) our results, which retain the lighting effects.} 
\label{fig:result_transfer}
\end{figure*}

\begin{figure}[t]
    \centering
    \includegraphics[width=\linewidth]{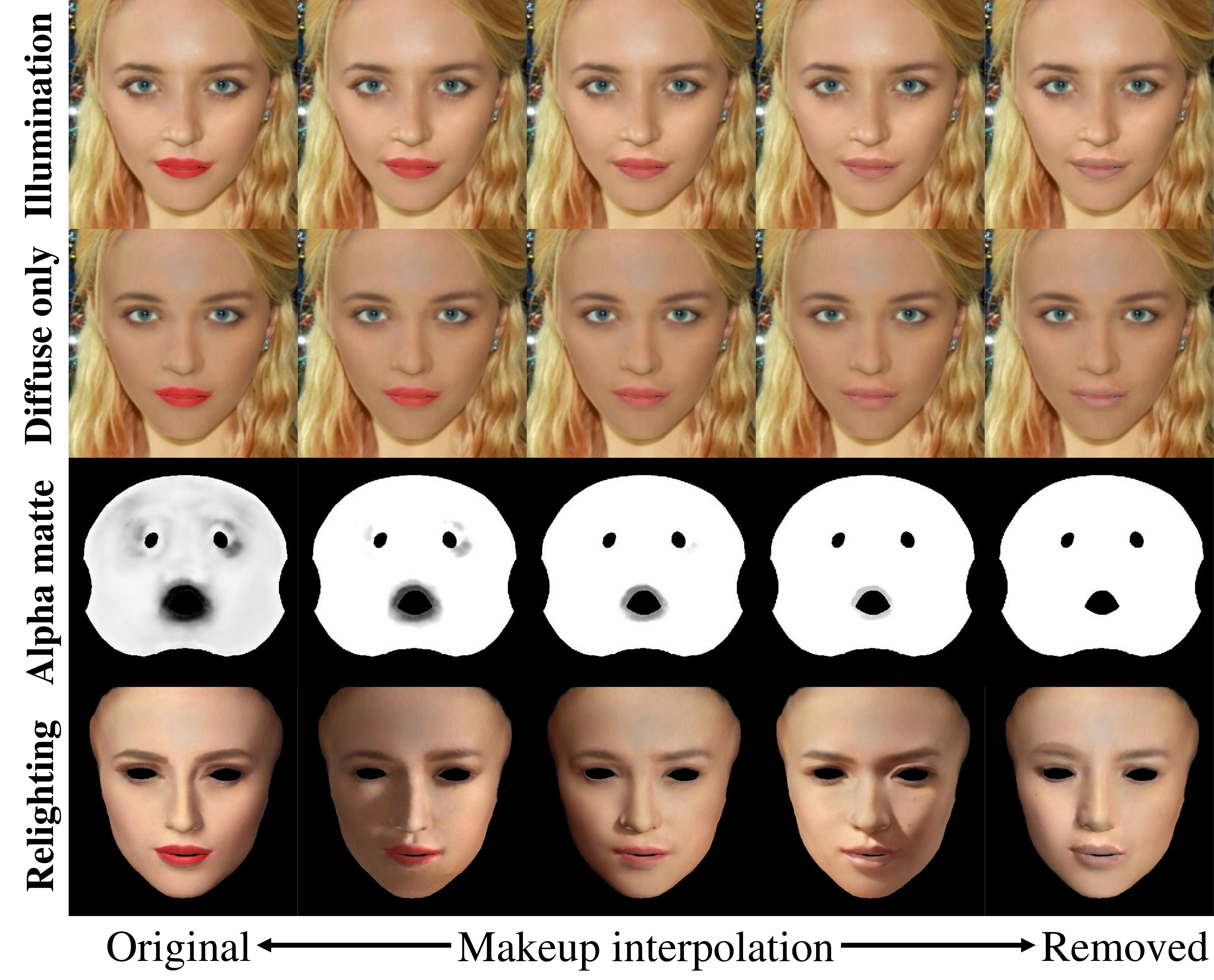}
    \caption{Results of illumination-aware makeup interpolation and removal. Left to right: the original makeup was interpolated and finally removed. 
    The first row presents the interpolation results with constant illumination. 
    The second row displays the results using only diffuse shading. The third row shows the alpha matte $\mathbf{A}$ for controlling the balance of the bare skin and makeup. The final row presents the rendered faces with relighting.} 
\label{fig:result_interpolate}
\end{figure}

\section{Applications}
\label{sec:application}

We explore several makeup-related applications using the results of our method and compare them with state-of-the-art methods.

\subsection{3D Face Reconstruction of Makeup}
First, we demonstrate how the extracted makeup dataset can be used to enhance the 3D face reconstruction of makeup. The same process as the diffuse albedo model construction of FLAME~\cite{FLAME:SiggraphAsia2017} was followed, and the collected makeup textures $(1 - \mathbf{A}) \odot \mathbf{D}_{m}^{m}$ were used to construct a PCA-based statistical model of makeup. The randomly sampled textures from the makeup model are depicted in Fig.~\ref{fig:result_pca_layer}. The makeup model is an extension of the diffuse albedo model, and the new model $\mathbf{D}_{c}^{'}$ can be formulated as:
\begin{align}
\mathbf{D}_{c}^{'} = \mathbf{D}_{c} + \mathbf{D}_{c}^{m} ,
\end{align}
where $\mathbf{D}_{c}^{m}$ is the makeup model.
We used an optimization-based manner to reconstruct the 3D face from the makeup portraits because no large-scale makeup dataset is available to train a neural network. A comparison of the 3D makeup face reconstruction using different albedo models $\mathbf{D}_{c}$ and $\mathbf{D}_{c}^{'}$ is presented in Fig.~\ref{fig:result_pca}. $\mathbf{D}_{c}^{'}$ could recover the makeup, and improve the accuracy of the entire reconstruction, especially for lipsticks and eye shadows. The shape of the lips and eyes were also matched more effectively by extending the ability of the diffuse albedo.

We believe that this makeup database can be explored further; for example, by using advanced image generation techniques such as StyleGAN/StyleGAN2~\cite{FFHQ, StyleGAN2} or diffusion model~\cite{diffusion}. The accuracy of the reconstruction results also requires further quantitative evaluation, and we consider these as future research topics.

\subsection{Illumination-Aware Makeup Transfer} 

We used the extracted makeup for makeup transfer by employing the same method as that for the makeup extraction network. Equation (\ref{eq:alhpa_blend}) was followed to blend a new makeup face, which was subsequently projected onto the original image.
The functionalities of our method and previous methods are summarized in Tab.~\ref{tab:table_function}. Similar to the approach of CPM~\cite{m_Nguyen-etal-CVPR21}, the makeup textures are UV representations to solve the misalignment of faces. The face region that is used for transfer can be specified, and the makeup is editable. In addition to these advantages, our approach extracts makeup and uses a completed UV representation, which can handle occlusion and illumination.

Our makeup transfer results are depicted in 
Fig.~\ref{fig:result_complex_transfer}. The reference images contained various complex factors, including occlusion, face misalignment, and lighting conditions. We extracted the makeup from the reference image. Subsequently, we transferred the makeup to the source image while maintaining the original illumination of the source image.

A qualitative comparison with state-of-art makeup transfer methods is depicted in Fig.~\ref{fig:result_transfer}, in which the source and reference images have different illumination. As existing makeup transfer methods do not consider illumination, they transfer not only the makeup, but also the effect of the illumination. The first two rows present the exchanged makeup results of two identities, one was evenly illuminated, whereas the other had shadows on the cheeks. The existing methods could not retain the illumination from the source image, which resulted in a mismatch between the face and surrounding environment. Note that our method preserved the illumination and shade of the original images in the cheeks. 
The final row shows an example in which the source and reference images have contrasting illumination. Our method enabled a natural makeup transfer, whereas the other methods were affected by the illumination, resulting in undesirable results. Note that the pioneering method known as BeautyGAN~\cite{Li:2018:MM} (a) was relatively stable. However, as it is not suitable for transferring eye shadow, the makeup in the source image was not cleaned up.

\subsection{Illumination-Aware Makeup Interpolation and Removal} 

As illustrated in Fig.~\ref{fig:result_interpolate}, compared to existing makeup transfer methods, our method could achieve makeup interpolation and removal without a reference image. Moreover, our makeup interpolation maintained constant illumination conditions and achieved natural makeup interpolation. The first two rows present the interpolation and removal results of the face images. The specular and diffuse shading were not changed while the makeup was adjusted from heavy to light. The third row shows how the alpha matte changed in the makeup interpolation process. 
The alpha matte $\mathbf{A}$ is adjusted to $\mathbf{A}_{\sigma}$ as follows:
\begin{align}
\mathbf{A}_{\sigma}(p) = clamp(\mathbf{A}(p) + \sigma, 0, 1) , \label{eq:alhpa_lerp}
\end{align}

where $\mathbf{A}_\sigma(p)$ and $\mathbf{A}(p)$ are values of $\mathbf{A}_\sigma$ and $\mathbf{A}$ at pixel $p$, respectively, and $\sigma \in [0, 1]$.
$\mathbf{A}_{\sigma}$ was eventually clipped to between 0 and 1. It can be observed that for the change in $\mathbf{A}_{\sigma}$, the makeup was mainly lipstick and eye shadow in this sample. Thus, the original $\mathbf{A}$ was the black color around the lips and eyes, which means that this part of the skin color was not used, while the makeup is mainly applied. With the increase in $\\sigma$, $\mathbf{A}_{\sigma}$ became whiter; thus the use of makeup was reduced and makeup interpolation was achieved. As the illumination was disentangled from the makeup, it was possible to relight the face while adjusting the makeup. The results are presented in the final row.

In addition to the above applications, the extracted makeup can be used for makeup recommendations, please refer to~\cite{Scherbaum11Makeup, 10.5555/3298239.3298377}. The makeup textures can also facilitate the subsequent processing of traditional graphics pipelines such as physically-based makeup rendering.

~\section{Limitations and Conclusions}
\label{sec:conclution}

\begin{figure}[t]
    \centering
    \includegraphics[width=0.75\linewidth]{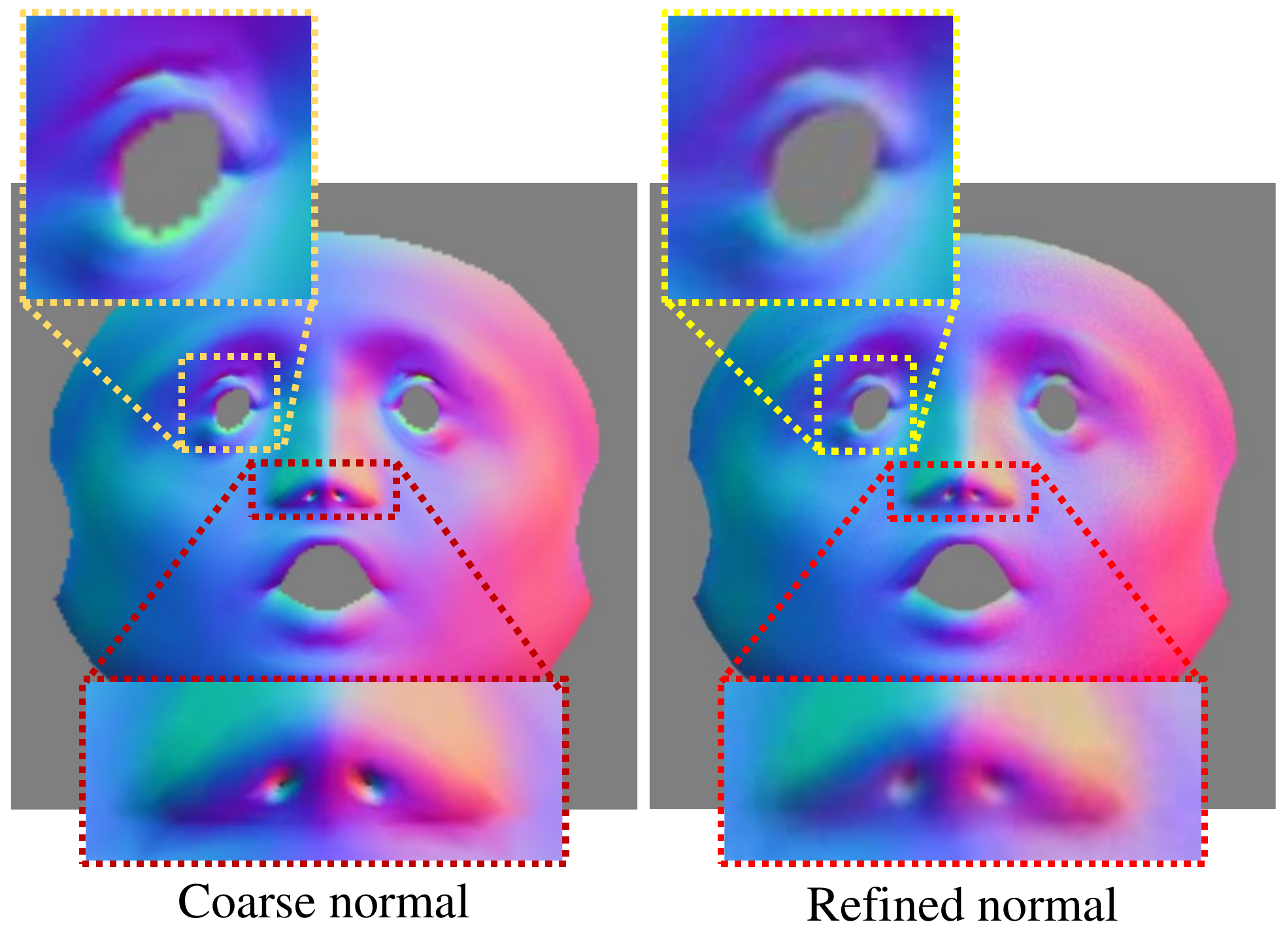}
    \caption{Comparison of coarse and refined normal.}
\label{fig:result_normal}
\end{figure}

\noindent\textbf{Limitations.} 
Although we integrated a color adjustment to alleviate the inherent skin color bias, the 3DMM skin colors still exhibited a problem. Namely, because the skin colors of the FLAME model~\cite{FLAME:SiggraphAsia2017} were obtained by unwrapping face images of the FFHQ dataset~\cite{FFHQ},
they contained baked-in lighting effects.
Therefore, our coarse albedos that were obtained using the FLAME model contained shading effects, which further caused errors in our refinement step and makeup extraction.
Whereas our method yielded satisfactory results in most cases, we would like to explore a better albedo model.

Our network erroneously extracts a makeup-like material from a makeup-less face because our network is currently not trained with paired data of makeup-less inputs and makeup-less outputs.

Although our refinement step greatly improves the diffuse albedos (see Fig.~\ref{fig:result_albedo}(b)(d)), the difference in the face geometry is subtle before and after refinement. Fig.~\ref{fig:result_normal} shows that the normals around the eyes and mouth were smoothed. We would like to improve the face geometry as well.

At present, we extract makeup from diffuse albedos, but in the real world, makeup contains specular albedos. We would like to account for makeup BRDFs for a more physically-plausible makeup transfer.
Quantitative evaluation is difficult in makeup-related research because no public ground-truth dataset of accurately aligned face images before and after makeups is available.
Therefore, it is also essential to establish a quantitative evaluation criterion.

\noindent\textbf{Conclusions.} 
We have presented the first method for extracting makeup for 3D face models from a single makeup portrait, which consists of the following three steps; 1) the extraction of coarse facial materials such as geometry and diffuse/specular albedos via extended regression-based inverse rendering using 3DMM~\cite{FLAME:SiggraphAsia2017}, 2) a newly designed optimization-based refinement of the coarse materials, and 3) a novel network that is designed for extracting makeup. Thanks to the disentangled outputs, we can achieve novel applications such as illumination-aware (i.e., relightable) makeup transfer, interpolation, and removal. The resultant makeup is well aligned in the UV space, from which we built a large-scale makeup texture dataset and a PCA-based makeup model. In future work, we would like to overcome the current limitations and explore better statistical models for facial makeup.
~\section*{Acknowledgements}
We thank Prof. Yuki Endo in University of Tsukuba for the suggestions and comments on our research. We also thank the reviewers for their constructive feedback and suggestions, which helped us improve our paper.

\bibliographystyle{eg-alpha-doi}
\bibliography{EG2023-body}

\end{document}